\documentclass[letterpaper, 10 pt, conference]{ieeeconf}

\usepackage{gensymb}

\makeatletter
\newcommand\notsotiny{\@setfontsize\notsotiny\@vipt\@viipt}
\makeatother

\usepackage{amsfonts}
\usepackage{amsmath}
\usepackage{amssymb}
\usepackage{balance}
\usepackage{blindtext}
\usepackage{centernot}
\usepackage{array}
\usepackage{graphicx}
\usepackage{adjustbox}
\usepackage{hyperref}
\usepackage[latin1]{inputenc}
\usepackage{lineno}
\usepackage[ruled,vlined,linesnumbered]{algorithm2e}
\usepackage{algpseudocode}

\usepackage{mathrsfs}
\usepackage{mathtools}
\usepackage{todonotes}
\usepackage{verbatim}
\usepackage{float}
\usepackage{upgreek}
\usepackage{mathdots}
\usepackage{listings}

\usepackage{times}
\usepackage{titlesec}

\let\labelindent\relax
\usepackage{enumitem}

\usepackage{dirtytalk}
\usepackage{wrapfig}

\usepackage[backend=biber,style=numeric-comp,sorting=none,maxbibnames=99]{biblatex}

\newcommand{\cost}{\text{cost}}

\newcommand{\costnew}{\cost'}
\newcommand{\costold}{\cost}
\newcommand{\dEKF}{dEKF}

\newcommand{\df}{d_f}
\newcommand{\dx}{d_x}
\newcommand{\dmu}{d_\mu}
\newcommand{\dz}{d_z}

\newcommand{\embed}{\text{embed}}

\newcommand{\fd}{f^{(m)}}
\newcommand{\fs}{f^{(s)}}
\newcommand{\fpose}{\xi}
\newcommand{\go}{g^o}

\newcommand{\Gfo}{G_f^o}
\newcommand{\Gxio}{G_{\xi}^o}
\newcommand{\hs}{h}

\newcommand{\invg}{\gamma}
\newcommand{\Invg}{\Gamma}

\newcommand{\mcalM}{\mathcal{M}}
\newcommand{\mcalN}{\mathcal{N}}

\newcommand{\N}{\mathbb{N}}
\newcommand{\nf}{n_f}
\newcommand{\Nf}{N_f}

\newcommand{\no}{n_o}
\newcommand{\nof}{n_{of}}

\newcommand{\proj}{\text{proj}}

\newcommand{\ra}{\rightarrow}
\newcommand{\R}{\mathbb{R}}
\newcommand{\Ra}{\Rightarrow}

\newcommand{\xr}{x}
\newcommand{\xrt}{x_t}

\newcommand{\zd}{z^{(m)}}
\newcommand{\zs}{z^{(s)}}

\newtheorem{theorem}{Theorem}[section]

\bibliography{references}

\begin{document}

\IEEEoverridecommandlockouts
\overrideIEEEmargins


\title{\LARGE \bf SLAM Backends with Objects in Motion: A Unifying Framework and Tutorial}

  \author{
      Chih-Yuan Chiu$^\star$
  \thanks{$^\star$Corresponding author. The author is with the EECS Department at the University of California, Berkeley, CA 94720 USA (email: \texttt{
  chihyuan\_chiu
  at berkeley dot edu.})
  }
  }  
  



\maketitle

\thispagestyle{empty}
\pagestyle{empty}

\allowdisplaybreaks

\begin{abstract}

Simultaneous Localization and Mapping (SLAM) algorithms are frequently deployed to support a wide range of robotics applications, such as autonomous navigation in unknown environments, and scene mapping in virtual reality. Many of these applications require autonomous agents to perform SLAM in highly dynamic scenes. To this end, this tutorial extends a recently introduced, unifying optimization-based SLAM backend framework to environments with moving objects and features \cite{Saxena2022SLAMNonlinearOptimization}. Using this framework, we consider a rapprochement of recent advances in dynamic SLAM. Moreover, we present \textit{dynamic EKF SLAM}: a novel, filtering-based dynamic SLAM algorithm generated from our framework, and prove that it is mathematically equivalent to a direct extension of the classical EKF SLAM algorithm to the dynamic environment setting. Empirical results with simulated data indicate that dynamic EKF SLAM can achieve high localization and mobile object pose estimation accuracy, as well as high map precision, with high efficiency.

\end{abstract}




\section{Introduction}
\label{sec: Introduction}

Simultaneous Localization and Mapping (SLAM) is a well-studied robotics problem in which an autonomous agent attempts to locate itself in an uncharted environment while constructing a map of said environment \cite{LeonardDurrantWhyte1991SimultaneousMapBuilding, Cadena2016PastPresentandFuture}. Most state-of-the-art SLAM algorithms operate under the \textit{static world} setting, in which the locations of landmarks in the robot's environment are assumed to be fixed. This greatly restricts the applicability of SLAM algorithms to robotics tasks such as autonomous navigation, in which SLAM-constructed maps must describe a wide variety of dynamic objects, such as moving obstacles, human-operated vehicles, or other autonomous agents.

To bridge this gap, the rapidly maturing \textit{dynamic SLAM} community aims to design SLAM algorithms that track moving objects while performing SLAM on the underlying static scene. Specifically, dynamic SLAM algorithms simultaneously estimate \textit{ego robot states}, \textit{static features}, \textit{features on moving objects}, and \textit{poses of moving objects}. To this end, Wang et al. proposed the SLAMMOT algorithm, which separately performs motion tracking for dynamic objects and SLAM over an underlying, fixed background \cite{Wang2007SLAMMOT}. Yang et al. introduced CubeSLAM, which assigns each dynamic object a rectangular bounding box, and tracks the boxes' trajectories across time \cite{Yang2019CubeSLAM}. Huang et al. proposed ClusterSLAM, which aggregates feature points corresponding to various dynamic objects in the scene, then performs bundle adjustment over each cluster \cite{Huang2019ClusterSLAM}. Bescos et al. presented DynaSLAM and DynaSLAM II, which uses the ORB-SLAM algorithm to extract features of, and subsequently track, dynamic objects \cite{bescos2018dynaslam, Bescos2021DynaSLAMII}. Zhang et al. introduce VDO-SLAM, which fuses dense optical flow and image segmentation to perform joint inference over robot poses, static landmark positions, and the pose and feature positions of mobile objects \cite{zhang2020VDOSLAM}. Although these approaches obtain reasonable accuracy in tracking moving objects, they typically incur a computational burden that increases rapidly with the number of moving objects tracked, and the length of the time horizon over which inference is performed. 

In this work, we extend the unifying, optimization-based SLAM formulation in \cite{Saxena2022SLAMNonlinearOptimization} to the dynamic SLAM setting. We illustrate that the aforementioned dynamic SLAM algorithms employ back-ends corresponding to different design choices in the context of our framework. To address the computational limitations of existing methods, we use our framework to derive \textit{dynamic EKF-SLAM}, a filtering-based algorithm that establishes a rapprochement between two classes of algorithms: efficient conventional filtering-based methods for static-world SLAM \cite{Thrun2005ProbabilisticRobotics}, and accurate but computationally costly bundle adjustment methods underlying existing dynamic SLAM algorithms. We prove that dynamic EKF SLAM is mathematically equivalent to a straightforward extension of the conventional EKF-SLAM algorithm to dynamic scenes. We then illustrate the empirical success of dynamic EKF-SLAM in performing inference over a simulated driving scenario, in which an ego autonomous vehicle travels down a highway in the presence of two other vehicles and a jaywalking pedestrian.

\section{Dynamic SLAM: A Unifying Framework}
\label{sec: Dynamic SLAM: A Unifying Framework}

Suppose that, at time $t \geq 0$, the estimated variables of the ego robot describe its past and/or present poses, $\nf \in \N$ static features, and $\no \in \N$ moving objects with $\nof(\alpha) \in \N$ features for each object index $\alpha \in \{1, \cdots, \no\}$. Given $n, n_1, n_2 \in \N$, with $n_1 < n_2$, set $[n] := \{1, \cdots, n\}$ and $[n_1: n_2] := \{n_1, \cdots, n_2 \}$. We have:
\begin{itemize}
    \item $\{\xr_t \in \R^{\dx}: t \in [T]\}$ denotes ego robot states, e.g., its poses and velocities, etc., relative to a global frame \textbf{G}.
    
    \item $\{\fs_k \in \R^{\df}: k \in [\nf] \}$ describes the current position estimate of each of the $\nf$ currently tracked static features relative to frame \textbf{G}, with corresponding feature measurements $\{\zs_{t, k} \in \R^{\dz}: t \in \{0\} \cup [T], k \in [\nf] \}$ at each time $t \in \{0\} \cup [T]$.
    
    \item $\{\fd_{t, \alpha, k} \in \R^{\df}: t \in \{0\} \cup [T], \alpha \in [\no], k \in [\nof(\alpha)] \}$ describes the feature position estimates, at each time $t \in \{0\} \cup [T]$, of each of the $k$ features on the $\alpha$-th moving object, maintained in the estimation window relative to frame \textbf{G}, with corresponding feature measurements $\{\zd_{t, \alpha, k} \in \R^{\dz}: t \in \{0\} \cup [T], \alpha \in [\no], k \in [\nof(\alpha)] \}$.
    
    \item $\{\fpose_{t, \alpha}: t \in [T], \alpha \in [\no] \}$ describes the poses of each of the $\no$ currently tracked moving objects at time $t$ relative to its pose at time $0$.
\end{itemize}


The evolution of states, features (associated with both static and moving objects), and moving object poses are captured by the following infinitely continuously differentiable (i.e., $C^\infty$) maps. The ego robot dynamics map $g: \R^{d_x} \ra \R^{d_x}$, the feature measurement map $h: \R^{d_x} \times \R^{d_f} \ra \R^{d_z}$, and the moving object pose transform map $\go: \R^{d_x} \times \R^{d_x} \ra \R^{d_x}$, are defined via additive noise models as shown below:
\begin{align} \label{Eqn: Ego Robot Dynamics Map}
    \xr_{t+1} &= g(\xrt) + w_t, \hspace{5mm} w_t \sim \mathcal{N}(0, \Sigma_w), 
    \\
    \label{Eqn: Measurement Map}
    z_{t,k} &= \hs(\xrt, f_k) + v_{t,k}, \hspace{5mm} v_{t,k} \sim \mathcal{N}(0, \Sigma_v), 
    \\
    \label{Eqn: Moving Object Pose Transform Map}
    \fd_{t, \alpha, k} &= \go(\fpose_{t, \alpha}, \fd_{0, \alpha, k}) + n_{t, \alpha}, \hspace{5mm} n_{t, \alpha} \sim \mathcal{N}(0, \Sigma_\fpose), \\ \nonumber
    &\hspace{5mm} \forall \hspace{0.5mm} t \in \{0\} \cup [T], \alpha \in [\no], k \in [\nof(\alpha)].
\end{align}
where $\mathcal{N}(\mu, \Sigma)$ denotes the Gaussian distribution with mean $\mu \in \R^d$ and covariance matrix $\Sigma \in \R^{d \times d}$, for some $d \in \N$, and $\Sigma_w \in \R^{\dx \times \dx}$, $\Sigma_v \in \R^{\dz \times \dz}$, $\Sigma_\fpose \in \R^{\df \times \df}$ are symmetric positive definite (p.d.) noise covariances. 
In the sections below, we assume that $\frac{\partial h}{\partial f_k}(x_t, f_k)$ is surjective at each $(x_t, f_k) \in \R^{\dx} \times \R^{\df}$, and that $\frac{\partial g^o}{\partial \fpose_{t, \alpha}}$ is injective at each $(\fpose_{t, \alpha}, \fd_{0, \alpha}) \in \R^{\dx} \times \R^{\df}$.



Our optimization-based formulation of dynamic SLAM includes the following steps, each of which updates the running cost term (\say{$\costold \ra \costnew$}). 

\begin{enumerate}
    \item \textbf{Feature Augmentation:}
    
    $\hspace{5mm}$ Let $\{z_k: k \in I_f\} \subset \R^{\dz}$ denote feature measurements, taken with respect to previously untracked features $\{f_k: k \in I_f\} \subset \R^{\df}$
    . 
    These may correspond to static or moving objects.
    The \textit{feature augmentation} step updates the running cost to include residual terms concerning these newly observed features:
    \begin{align*}
        \costnew = \costold + \sum_{k \in I_f} \Vert z_{t,k} - \hs(\xrt, f_k) \Vert_{\Sigma_v^{-1}}^2
    \end{align*}
    
    \item \textbf{Moving Object Pose Augmentation:}
    
    $\hspace{5mm}$ Let $\{\fd_{t, \alpha, k} : k \in I_{f, \alpha}, \alpha \in I_o \}$ denote features of tracked moving objects that have been observed at times $t_1$ and $t_2$, with $t_1 < t_2$. For simplicity, define:
    \begin{align*}
        \fd_{\tau, \alpha} &:= (\fd_{\tau, \alpha, 1}, \cdots, \fd_{\tau, \alpha, \nof(\alpha)}) \in \R^{\nof(\alpha) \df},
    \end{align*}
    for each $\alpha \in [\no]$, $\tau \in \{0, t\}$.
    The \textit{moving object pose augmentation} step appends the current pose estimates of tracked moving objects, i.e., $\{\fpose_{t, \alpha}: \alpha \in [\no]\}$, to the running cost:
    \begin{align*}
        \costnew = \costold + \sum_{\alpha \in I_o} \sum_{k \in I_f, \alpha} \Vert \fd_{t, \alpha, k} - \go \big(\fpose_{t, \alpha}, \fd_{0, \alpha, k} \big) \Vert_{\Sigma_\fpose^{-1}}^2
    \end{align*}
    
    \item \textbf{Static Feature Update:}
    
    $\hspace{5mm}$ Let $\{z_k: k \in I_f\} \subset \R^{\dz}$ denote feature measurements, taken with respect to previously tracked static features $\{\fs_k: k \in I_f\} \subset \R^{\df}$
    . The \textit{feature update} step updates the cost as follows:
    \begin{align*}
        \costnew = \cost + \sum_{k \in I_f} \Vert \zs_{t,k} - \hs(\xrt, \fs_k) \Vert_{\Sigma_v^{-1}}^2
    \end{align*}
    
    \item \textbf{Smoothing Factor Augmentation:}
    
    $\hspace{5mm}$ The \textit{smoothing factor augmentation} step constrains the most recent moving object pose transformation ($\fpose_{t-1, \alpha}$ to $\fpose_{t, \alpha}$) from significantly differing from the second most recent moving object pose transformation ($\fpose_{t-2, \alpha}$ to $\fpose_{t-1, \alpha}$), for each object indexed $\alpha \in [\no]$:
    \begin{align*}
        \costnew = \cost + \sum_{\alpha \in I_o} \Vert s(\fpose_{t-2,\alpha}, \fpose_{t-1,\alpha}, \fpose_{t,\alpha}) \Vert_{\Sigma_s^{-1}}^2
    \end{align*}
    Here, $s: \R^{3 \dx} \ra \R^{\dx}$ is a smoothing function, e.g., for $\dx = 1$, take $s(\xi_{t-2, \alpha}, \xi_{t-1, \alpha}, \xi_{t, \alpha}) := (\xi_{t, \alpha} - \xi_{t-1, \alpha}) - (\xi_{t-1, \alpha} - \xi_{t-2, \alpha})$.
    
    \item \textbf{State Propagation:}
    
    $\hspace{5mm}$ At each time $t$, the \textit{state propagation step} updates the cost to include residual terms involving the odometry measurements between $\xrt$, the pose at time $t$, and $\xr_{t+1}$, the pose at time $t+1$:
    \begin{align*}
        \costnew = \cost + \Vert \xr_{t+1} - g(\xrt) \Vert_{\Sigma_w^{-1}}^2
    \end{align*}
    
\end{enumerate}

Poses and features present in the optimization window may be dropped (instead of marginalized) to improve optimization accuracy, as is common in SLAM algorithms operating under the static world assumption \cite{MurArtal2017ORBSLAM2, Leutenegger2015KeyframebasedVO}. In addition, the above formulation naturally extends to scenarios in which dynamical quantities evolve on smooth manifolds, rather than on Euclidean spaces (see \cite{Saxena2022SLAMNonlinearOptimization}, Section 3 and Appendix A).

\section{Unifying Existing Algorithms}
\label{sec: Unifying Existing Algorithms}

In this section, we interpret the back-ends of recently proposed dynamic SLAM algorithms as the selection of different design choices within the context of our framework, as presented in Section \ref{sec: Dynamic SLAM: A Unifying Framework}. We focus in particular on design choices relevant to tracking moving objects.

\begin{itemize}
    \item \textbf{CubeSLAM} \cite{Yang2019CubeSLAM}---In CubeSLAM, pose estimates of moving objects are obtained by forming and tracking rectangular bounding boxes across time. Feature augmentation of moving objects into the estimation window is avoided.
    
    \item \textbf{ClusterSLAM} \cite{Huang2019ClusterSLAM}--- ClusterSLAM models moving objects by aggregating and tracking feature clouds. The authors describe \say{fully-coupled}, \say{semi-decoupled}, and \say{decoupled} estimation schemes for static SLAM and moving object tracking, which correspond to increasingly aggressive marginalization schemes in our framework. 
    
    \item \textbf{VDO-SLAM} \cite{zhang2020VDOSLAM}---The VDO-SLAM algorithm performs object segmentation, then samples dense feature clouds within each bounding box to track the associated moving object. In contrast with CubeSLAM, this is a vigorous feature and pose augmentation scheme, with little marginalization within the estimation window. VDO-SLAM can enjoy considerable accuracy, but may also incur high computational burden \cite{Bescos2021DynaSLAMII}.
    
    \item \textbf{DynaSLAM II} \cite{Bescos2021DynaSLAMII}--- DynaSLAM II tracks moving objects across time, by repeatedly performing pose augmentation with pose estimates constructed from newly observed features. Unlike VDO-SLAM, the most recent feature position estimates of these moving objects are then quickly dropped or marginalized, to reduce the computation burden at the next timestep.
\end{itemize}

\begin{algorithm} \label{Alg: Dynamic EKF, Optimization-Based}

{
\small
\SetAlgoLined


\KwData{Prior $\mathcal N(\mu_0, \Sigma_0)$ on $\xr_0 \in \R^{d_x}$, noise covariances $\Sigma_w \in \R^{\dx \times \dx}$, $\Sigma_v \in \R^{\dz \times \dz}$, $\Sigma_\fpose \in \R^{\df \times \df}$, $\Sigma_s \in \R^{\dx \times \dx}$, dynamics map $g: \R^{\dx } \ra \R^{\dx}$, measurement map $h: \R^{\dx} \times \R^{\df} \ra \R^{dz}$, inverse measurement map $\ell: \R^{\dx} \times \R^{\dz} \ra \R^{df}$, moving object dynamics map $\go: \R^{\dx} \times \R^{\df} \ra \R^{\df}$ for each object indexed $\alpha \in [\no]$, inverse moving object dynamics map $\invg^\alpha: \R^{\nof(\alpha) \df} \times \R^{\nof(\alpha) \df} \ra \R^{\dx}$, time horizon $T \in \N$, number of features $\nf \in \N$, number of moving objects $\no \in \N$.}

\KwResult{Estimates $\mu_t 
, \hspace{0.5mm} \forall \hspace{0.5mm} t \in \{1, \cdots, T\}$.}


\vspace{2mm}
 $\cost_0 \gets \|\xr_0 - \mu_0\|^2_{\Sigma_0^{-1}} $
 
 
 $\nf, \no \gets 0$.
 
 \vspace{2mm}
 \For{$t = 0, 1, \cdots T-1$}{
  

    $\nf \gets$ Number of tracked features on static features

    $\Nf \gets$ Total number of tracked features on moving objects
    
    $\{z_{t,k}: k \in [\nf+\Nf+1: \nf+\Nf+\Nf'] \} \gets$ Measurements of new features, corresponding to both static landmarks and moving objects.
    
    $\cost_t \gets \cost_t + \sum_{k=\nf+\Nf+1}^{\nf+\Nf+\nf'+\Nf'} \|\zs_{t,k} - \hs(\xrt, \fs_k)\|_{\Sigma_{v}^{-1}}^2$.
        
    $\mu_t \gets \big(\mu_t, \ell(\xrt, \zs_{t,\nf+\Nf+1}), \cdots, \ell(\xrt, \zs_{t,\nf+\Nf+\nf'+\Nf'}) \big) 
    $.
        
    $\mu_t, \Sigma_t \gets$ Gauss-Newton, on $\cost_t$, about $\mu_t$ (\cite{Saxena2022SLAMNonlinearOptimization}, Alg. 3).

    Increment $\nf$, $\no$, $\{\nof(\alpha): \alpha \in [\no]\}$ as appropriate, given the newly detected $\nf'$ static features and $\Nf'$ features on moving objects.

    \If{$\no \geq 1$}{

        $\cost_t \gets \cost_t + \sum_{\alpha=1}^{\no} \sum_{k=1}^{\nof(\alpha)} \Vert \fd_{t, \alpha, k} - \go(\fpose_{t, \alpha}, \fd_{t, \alpha, k}) \Vert_{\Sigma_\fpose^{-1}}^2 \cdot \textbf{1}\{\fd_{t, \alpha, k} \text{ defined} \}$.
        
        $\mu_t \gets \big( \mu_t, \invg^\alpha \big(\fd_{0, 1}, \fd_{t, 1} \big), \cdots, \invg^\alpha \big(\fd_{0, \no}, \fd_{t, \no} \big) \big).$
            
        $\mu_t , \Sigma_t \gets$ Gauss-Newton, on $\cost_t$, about $\mu_t$ (\cite{Saxena2022SLAMNonlinearOptimization}, Alg. 3).
        
        (Optional) Drop $\{\fd_{t, \alpha}: \alpha \in [\no] \}$ from the mean and covariance estimates.
        
        
        
        
    }
    
    $\cost_t \gets \cost_t + \sum_{\alpha=1}^{\no} \Vert s(\fpose_{t-2, \alpha}, \fpose_{t, \alpha}, \fpose_{t, \alpha}) \Vert_{\Sigma_s^{-1}}^2 \cdot \textbf{1}\{\fpose_{t-2, \alpha}, \fpose_{t, \alpha} \text{ defined}\}$
    
    $\mu_t , \Sigma_t \gets$ Gauss-Newton, on $\cost_t$, about $\mu_t$ (\cite{Saxena2022SLAMNonlinearOptimization}, Alg. 3).

    $\{\zs_{t,k}: k \in [\nf]\} \gets$ Measurements of existing static features.
  
  $\cost_t \gets \cost_t + \sum_{k=1}^{\nf} \|\zs_{t,k} - \hs(\xrt, \fs_k)\|_{\Sigma_{v}^{-1}}^2$.
  
  $\bar\mu_t
  , \bar\Sigma_t 
   \gets$ Gauss-Newton, on $\cost_t$, about $\mu_t$,
  (\cite{Saxena2022SLAMNonlinearOptimization}, Alg. 3).
  
  

  $\cost_t \gets \cost_t + \|\xr_{t+1} - g(\xrt)\|_{\Sigma_w^{-1}}^2 $
  
  $\mu_{t+1} 
  , \Sigma_{t+1} 
   \gets$ Marginalization, on $\cost_{t+1}$ with $\xr_M = \xrt$,
  about $(\overline{\mu_t}, g(\overline{\mu_t}))$
  (\cite{Saxena2022SLAMNonlinearOptimization}, Alg. 4).
  
  $\cost_{t+1} \gets \|\xr_{t+1} - \mu_{t+1}\|^2_{\Sigma_{t+1}^{-1}} $
  
 }
 
 \Return{$\mu_0, \cdots, \mu_T$}
 \caption{Dynamic EKF SLAM, as Iterative optimization.}
 }
\end{algorithm}

\begin{algorithm} \label{Alg: Dynamic EKF, Standard}

{
\small
\SetAlgoLined


\KwData{Prior $\mathcal N(\mu_0, \Sigma_0)$ on $\xr_0 \in \R^{d_x}$, noise covariances $\Sigma_w \in \R^{\dx \times \dx}$, $\Sigma_v \in \R^{\dz \times \dz}$, $\Sigma_\fpose \in \R^{\df \times \df}$, $\Sigma_s \in \R^{\dx \times \dx}$, dynamics map $g: \R^{\dx } \ra \R^{\dx}$, measurement map $h: \R^{\dx} \times \R^{\df} \ra \R^{dz}$, inverse measurement map $\ell: \R^{\dx} \times \R^{\dz} \ra \R^{df}$, moving object dynamics map $\go: \R^{\dx} \times \R^{\df} \ra \R^{\df}$ for each object indexed $\alpha \in [\no]$, inverse moving object dynamics map $\invg^\alpha: \R^{\nof(\alpha) \df} \times \R^{\nof(\alpha) \df} \ra \R^{\dx}$, time horizon $T \in \N$, number of features $\nf \in \N$, number of moving objects $\no \in \N$.}

\KwResult{Estimates $\mu_t 
, \hspace{0.5mm} \forall \hspace{0.5mm} t \in \{0, 1, \cdots, T\}$.}

\vspace{2mm}
 $\cost_0 \gets \|\xr_0 - \mu_0\|^2_{\Sigma_0^{-1}} $
 
 
 $\nf, \no \gets 0$.
 
 \vspace{2mm}
 \For{$t = 0, 1, \cdots T-1$}{
  

    $\{\zs_{t,k}: k \in [\nf+1: \nf+\nf'] \} \gets$ Measurements of new static features.
    
    $\mu_t, \Sigma_t, \nf \gets$ \text{ Alg. \ref{Alg: Dynamic EKF, Feature Augmentation, Standard}, Dynamic EKF, (Static) Feature Augmentation}

        

        
    
    \If{$\no \geq 1$}{
        $\{\zd_{t,\alpha, k}: \alpha \in [\no], k \in [\nof(\alpha) + \nof(\alpha)'] \} \gets$ Measurements of $\nof(\alpha)$ tracked and $\nof(\alpha)'$ new features of previously tracked moving objects indexed $\alpha \in [\no]$.
        
        $\mu_t, \Sigma_t \gets$ \text{ Alg. \ref{Alg: Dynamic EKF, Feature Augmentation, Standard}, Dynamic EKF, (Dynamic)} \text{ Feature Augmentation}
    
        
        
        $\mu_t, \Sigma_t \gets$ \text{ Alg. \ref{Alg: Dynamic EKF, Moving Object Pose Augmentation, Standard}, Dynamic EKF, (Dynamic) Object} \text{ Pose Augmentation}

        
            
    
    }
    
    \If{\emph{detect $n_o' \geq 1$ new moving objects}}{
        $\{\zd_{t,\alpha, k}: \alpha \in [\no+1: \no + \no'], k \in [\nof(\alpha)] \} \gets$ Measurements of features of new moving objects.
        
        $\mu_t, \Sigma_t \gets$ \text{ Alg. \ref{Alg: Dynamic EKF, Feature Augmentation, Standard}, Dynamic EKF, (Dynamic)} \text{Feature Augmentation}
        
        $\no \gets \no + \no'$.
        
        
        
    }
    
    $\mu_t, \Sigma_t \gets$ \text{ Alg. \ref{Alg: Dynamic EKF, Smoothing Update, Standard}, Dynamic EKF, Smoothing Update}
    
    

    $\{\zs_{t,k}: k \in [\nf]\} \gets$ Measurements of existing static features.
    
    $\overline{\mu}_t, \overline{\Sigma}_t \gets$ \text{ Alg. \ref{Alg: Dynamic EKF, Static Feature Update, Standard}, Dynamic EKF, Static Feature Update}
    
    $\mu_{t+1}, \Sigma_{t+1} \gets$ \text{ Alg. \ref{Alg: Dynamic EKF, State Propagation, Standard}, Dynamic EKF, State Propagation}

  
  
  

  
  
  
 }
 
 \Return{$\mu_0, \cdots, \mu_T$}
 \caption{Dynamic EKF SLAM, Standard formulation.}
 }
\end{algorithm}

\section{Dynamic EKF-SLAM}
\label{sec: Dynamic EKF-SLAM}

Although the algorithms described in \ref{sec: Unifying Existing Algorithms} can attain high estimation accuracy, their computation time often scales poorly with the number of moving objects or timesteps tracked. Inspired by the efficiency of filtering-based SLAM frameworks under the static world assumption, we use the unifying framework presented in Section \ref{sec: Dynamic SLAM: A Unifying Framework} to construct the dynamic EKF algorithm, described below, to address this issue.

At each time $t$, the dynamic EKF SLAM algorithm on Euclidean spaces maintains the full state vector:
\begin{align} \label{Eqn: Dynamic EKF SLAM, Full State}
    \tilde{x}_t = (\xrt, \fs, \fd, \fpose) \in \R^{\dmu}.
\end{align}
where $\dmu := \dx + \nf \df + 2 \cdot \sum_{\alpha=1}^{\no} \nof(\alpha) \df + (t-1) \no \dx$. (For generality, we assume that all past moving object poses are maintained; in practice, these can be dropped). The components of $\tilde x_t$ are as follows:    
\begin{itemize}
    \item \textbf{Ego robot pose}:
    
    $\hspace{5mm}$ $\xrt \in \R^{\dx}$ denotes the ego robot pose at the current time $t$.
    
    \item \textbf{Static feature position estimates}:
    
    $\hspace{5mm}$ $\fs := (\fs_1, \cdots, \fs_{\nf}) \in \R^{\nf \df}$ is the position estimates of the $\nf \in \N$ static features currently tracked.
    
    \item \textbf{Moving object feature position estimates}:
    
    $\hspace{5mm}$ $\fd$, defined below, is the feature positions of moving objects at the initial time $0$ and the current time $t$. Here, $\fd_{\tau, \alpha, k} \in \R^{\df}$ denotes the position estimate of the $k$-th feature of the moving object indexed $\alpha$ at time $\tau$, for each $\tau \in \{0, t\}$, $\alpha \in [\no]$, and $k \in [\nof(\alpha)]$, and $\Nf := \sum_{\alpha=1}^{\no} \nof(\alpha)$ denotes the total number of features summed over all moving objects:
    \begin{align*}
        &\hspace{8mm} \fd \\
        &:= \big( \fd_{0, 1, 1}, \cdots, \fd_{0, 1, \nof(1)}, 
        \cdots, \fd_{0, \no, 1}, \cdots, \fd_{0, 1, \nof(\no)}, \\
        &\hspace{5mm} \fd_{t, 1, 1}, \cdots, \fd_{t, 1, \nof(1)}, 
        \cdots, \fd_{t, \no, 1}, \cdots, \fd_{t, 1, \nof(\no)} \big) \\
        &\hspace{5mm} \in \R^{2 \cdot \Nf \cdot \df},
    \end{align*}
    
    $\hspace{5mm}$ For notational simplicity, we assume all features on all moving objects have been observed since the start of the time horizon. (This assumption can easily be relaxed). 
    
    \item \textbf{Moving object poses}:
    
    $\hspace{5mm}$ $\fpose := (\fpose_{1, 1}, \cdots, \fpose_{1, \no}, \cdots, \fpose_{t, 1}, \cdots, \fpose_{t, \no}) \in \R^{t \no \dx}$ denotes the past and present poses of the $\no$ objects currently tracked. Here, $\fpose_{\tau, \alpha} \in \R^{\dx}$ denotes the pose, of the moving object indexed $\alpha \in [\no]$, at time $\tau$, for each $\alpha \in [\no]$ and $\tau \in [t]$. To ensure computational tractability, past pose estimates may be dropped.
\end{itemize}

Below, if unspecified, we assume the components in the full state $\tilde x_t \in \R^{\do}$ appear in the order given in \eqref{Eqn: Dynamic EKF SLAM, Full State}, i.e., $\tilde x_t = (\xrt, \fs, \fd_t, \fpose_t) \in \R^{\dmu}$.

At initialization ($t = 0$), no feature or object has been detected ($\nf = \no = 0$, $\dmu = \dx$), and the dynamic EKF full state is simply the initial state $\tilde{x}_0 = \xr_0 \in \R^{\dx}$, with mean $\mu_0 \in \R^{\dx}$ and covariance $\Sigma_0 \in \R^{\dx \times \dx}$. 
Suppose, at some time $t$, the running cost $c_{\dEKF,t,0}: \R^{\dmu} \ra \R^{\dmu}$ is:
\begin{align*}
    c_{\dEKF,t,0} = \Vert \tilde{x}_t - \mu_t \Vert_{\Sigma_t^{-1}}^2,
\end{align*}
where $\tilde{x}_t \in \R^{\dmu}$ denotes the EKF full state at time $t$, as described in the paragraphs above, with mean $\mu_t \in \R^{\dmu}$ and symmetric positive definite covariance matrix $\Sigma_t \in \R^{\dmu \times \dmu}$.

Let $\Nf := \nf + \sum_{\alpha=1}^{\no} \nof(\alpha)$ denote the total number of features (static and moving) tracked at time $t$. First, the \textit{feature augmentation step} affixes new features' maximum a posteriori position estimates, denoted $f_{\nf+\Nf+1}, \cdots, f_{\nf+\Nf+\nf'+\Nf'} \in \R^{\df}$ to the EKF full state $\tilde{x}_t$, and updates the mean and covariance of the full state. These new features may belong to static landmarks, previously detected moving objects, or new, previously undetected moving objects. Feature measurements $z_{t, \nf+\Nf+1}, \cdots, z_{t, \nf+\Nf+\nf'+\Nf'} \in \R^{\dz}$ are incorporated by adding measurement residuals to the current running cost $c_{\dEKF,t,0}$, resulting in a new cost $c_{\dEKF,t,1}: \R^{\dmu + (\nf'+\Nf') d_f} \ra \R$:
\begin{align*}
    &c_{\dEKF,t,1}(\tilde{x}_t, f_{t,\nf + \Nf + 1}, \cdots, f_{t,\nf + \Nf + \Nf'}) \\
    := \hspace{0.5mm} &\Vert \tilde{x}_t-\mu_t \Vert_{\Sigma_t^{-1}}^2 + \sum_{k= \Nf+1}^{\Nf+\Nf'} \Vert z_{t,k} -\hs(\xrt, f_{t,k}) \Vert_{\Sigma_v^{-1}}^2.
\end{align*}
Thus, $c_{\dEKF,t,1}(\tilde{x}_t, f_{t,\nf+\Nf+1}, \cdots, f_{t,\nf+\Nf+\nf'+\Nf'})$ incorporates new feature positions to $\tilde{x}_t$, and constrains it using feature measurements residuals. A Gauss-Newton step then updates the mean $\mu_t \in \R^{\dmu + \Nf' \df}$ and covariance $\Sigma_t \in \R^{(\dmu + (\nf'+\Nf') \df) \times (\dmu + (\nf'+\Nf') \df)}$ for $\tilde x_t$, resulting in a new cost:
\begin{align*}
    c_{\dEKF,t,2}(\tilde{x}_t) 
    &:= \Vert \tilde{x}_t-\mu_t \Vert_{\Sigma_t^{-1}}^2.
\end{align*}
We then increase $\dmu$ by $\Nf' \df$, adjoin the new feature variables $(f_{t, \nf+\Nf+1}, \cdots, f_{t, \nf+\Nf+\nf'+\Nf'})$ to $\tilde x_t$, and rearrange the components of the full state $\tilde x_t$ so that those new features associated with previously detected moving objects are stored alongside previously detected features for the same object (as determined by data assocation in the front end). If some new features correspond to a newly detected object, we store those features together as adjacent components in $\tilde x_t$, and accordingly increment $\no$ (the number of objects currently stored inside $\tilde x_t$). This restores the full state $\tilde x_t$ to the form $(\xrt, \fs, \fd, \fpose) \in \R^{\dmu}$ in \eqref{Eqn: Dynamic EKF SLAM, Full State}.

The \textit{moving object pose augmentation} step then appends pose estimates, denoted $\fpose_{\alpha, t}$, for each tracked moving object $\alpha \in [\no]$, relative to their initial pose. Moving objects' pose residual terms are added to the current running cost $c_{\dEKF, t, 2}$, resulting in a new cost $c_{\dEKF, t, 3}: \R^{\dmu} \times \R^{\no \dx} \ra \R$:
\begin{align*}
    &c_{\dEKF, t, 3}(\tilde x_t, \fpose_{t, 1}, \cdots, \fpose_{t, \no}) \\
    = \hspace{0.5mm} & \Vert \tilde x_t - \mu_t \Vert_{\Sigma_t^{-1}}^2 + \Vert \fd_{t, \alpha} - \go \big(\fpose_{t, \alpha}, \fd_{t, \alpha} \big) \Vert_{\Sigma_\fpose^{-1}}^2.
\end{align*}
Essentially, $c_{\dEKF, t, 3}(\tilde x_t, \fpose_{t, 1}, \cdots, \fpose_{t, \no})$ appends positions of new moving object poses to the full state $\tilde x_t$, and constrains it using the pose transform map $\go: \R^{\dx} \times \R^{\df} \ra \R^{\df}$. A Gauss-Newton step then constructs an updated mean $\mu_t \in \R^{\dmu}$ and an updated covariance matrix $\Sigma_t \in \R^{\dmu \times \dmu}$ for $\tilde x_t$, resulting in a new cost $c_{\dEKF, t, 4}(\tilde x_t): \R^{\dmu + \no \dx} \ra \R$:
\begin{align*}
    c_{\dEKF, t, 4}(\tilde x_t) := \Vert \tilde x_t - \mu_t \Vert_{\Sigma_t^{-1}}^2.
\end{align*}
We adjoin the new moving object poses $(\fpose_{t, 1}, \cdots, \fpose_{t, \no})$ to $\tilde x_t$ (or record and drop them), then rearrange the components of the full state $\tilde x_t$ so that each new moving object pose is stored alongside previously tracked poses for the same object. This restores the full state $\tilde x_t$ to the form $(\xrt, \fs, \fd_m, \fpose_t) \in \R^{\dmu}$, as introduced previously in \eqref{Eqn: Dynamic EKF SLAM, Full State}.

Next, the \textit{static feature update} step uses measurements of features contained in $\tilde x_t$ to update the mean and covariance of $\tilde x_t$. More precisely, measurements $\zs_{t,1}, \cdots, \zs_{t,\nf} \in \R^{\dz}$, of the $\nf$ static features $\fs_1, \cdots, \fs_{\nf} \in \R^{\df}$ currently tracked in $\tilde{x}_t$, are introduced by incorporating associated measurement residuals 
to the running cost, resulting in a new cost $c_{\dEKF,t,5}: \R^{\dmu} \ra \R$:
\begin{align*}
    &c_{\dEKF,t,5}(\tilde{x}_t) \\
    := \hspace{0.5mm} &\Vert \tilde{x}_t-\mu_t \Vert_{\Sigma_t^{-1}}^2 + \sum_{k=1}^{\nf} \Vert z_{t,k} -\hs(\xrt, \fs_{t,k}) \Vert_{\Sigma_v^{-1}}^2.
\end{align*}
A Gauss-Newton step then constructs an updated mean $\mu_t \in \R^{\dmu}$ and covariance $\Sigma_t \in \R^{\dmu \times \dmu}$ for $\tilde{x}_t$, resulting in a new cost $c_{\dEKF,t,6}: \R^{\dmu} \ra \R$:
\begin{align*}
    c_{\dEKF,t,6}(\tilde{x}_t) 
    &:= \Vert \tilde{x}_t - \mu_t \Vert_{\Sigma_t^{-1}}^2,
\end{align*}
of the form of $c_{\dEKF,t,0}$.

The \textit{smoothing update} step then updates the three most recent tracked dynamic object poses, denoted $\{\fpose_{\tau, \alpha}: \tau \in \{t-2, t-1, t\}, \alpha \in [\no] \}$, by ensuring that the object's motion from time $t-2$ to time $t-1$ does not deviate significantly from its motion from time $t-1$ to time $t$. This regularization process ensures that the estimated trajectories of the moving objects are smooth enough to be physically feasible. To this end, we define a new cost $c_{\dEKF, t, 7}: \R^{\dmu} \ra \R$:
\begin{align*}
    &c_{\dEKF, t, 7}(\tilde x_t) \\
    := \hspace{0.5mm} &\Vert \tilde x_t - \mu_t \Vert_{\Sigma_t^{-1}}^2 + \sum_{\alpha=1}^{\no} \Vert s(\fpose_{t-2, \alpha}, \fpose_{t-1, \alpha}, \fpose_{t, \alpha}) \Vert_{\Sigma_s^{-1}}^2.
\end{align*}
We then apply a Gauss-Newton step to update the mean $\mu_t \in \R^{\dmu}$ and covariance $\Sigma_t \in \R^{\dmu \times \dmu}$, resulting in a new cost:
\begin{align*}
    c_{\dEKF, t, 8}(\tilde x_t) := \Vert \tilde x_t - \mu_t \Vert_{\Sigma_t^{-1}}^2.
\end{align*}
that has the form of the original cost $c_{\dEKF,t,0}$.

Finally, the \textit{state propagation} step advances the EKF full state forward in time, via the EKF state propagation map $g: \R^{\dmu} \ra \R^{\dmu}$. To pass $\tilde{x}_t$ forward to $\tilde{x}_{t+1}$, we absorb the dynamics residual into the running cost, resulting in a new cost $c_{\dEKF,t,9}: \R^{\dmu} \ra \R$:
\begin{align*}
    &c_{\dEKF,t,9}(\tilde{\xrt}, \xr_{t+1}) \\
    := \hspace{0.5mm} &\Vert \tilde{x}_t- \mu_t \Vert_{\Sigma_t^{-1}}^2 + \Vert \xr_{t+1} - g(\xrt) \Vert_{\Sigma_w^{-1}}^2,
\end{align*}
i.e., $c_{\dEKF,t,9}$ appends the new state $\xr_{t+1} \in \R^{\dx}$ to $\tilde{x}_t$, while adding a new cost encoded by the dynamics residuals. The algorithm then applies a marginalization step, with $\tilde{x}_{t,K} := (\xr_{t+1}, \fs, \fd, \fpose) \in \R^{\dmu}$ and $\tilde{x}_{t,M} := \xrt \in \R^{\dx}$, to remove the previous state $\xrt \in \R^{\dx}$ from the running cost. This step produces a mean $\mu_{t+1} \in \R^{\dmu}$ and a covariance $\Sigma_{t+1} \in \R^{\dmu \times \dmu}$ for the new EKF full state, $\tilde{x}_{t+1} := \tilde{x}_{t,K} = (\xr_{t+1}, \fs, \fd, \fpose)$. The running cost is updated to $c_{\dEKF,t+1,0}: \R^{\dmu} \ra \R$, defined by:
\begin{align*}
    c_{\dEKF,t+1,0}(\tilde{x}_{t+1}) 
    &:= \Vert \tilde{x}_{t+1} - \mu_{t+1} \Vert_{\Sigma_{t+1}^{-1}}^2,
\end{align*}
which assumes the form of $c_{\dEKF,t,0}$.

The theorems below establish the mathematical equivalence of the five steps of the dynamic EKF, as presented above in our optimization framework (Alg. \ref{Alg: Dynamic EKF, Optimization-Based}), to those presented in the extension standard EKF SLAM algorithm to a dynamic setting (Alg. \ref{Alg: Dynamic EKF, Standard}).
Theorem statements and proofs concerning the equivalence of the feature augmentation, feature update, and state propagation steps are identical to those in the static EKF-SLAM case, and are omitted for brevity. For more details, please see Appendix \ref{subsec: A1, Proofs of Main Theorems}, or \cite{Saxena2022SLAMNonlinearOptimization}, Theorems 5.1-5.3.

\begin{figure*}[!htbp]
    \centering
    \includegraphics[scale=0.2]{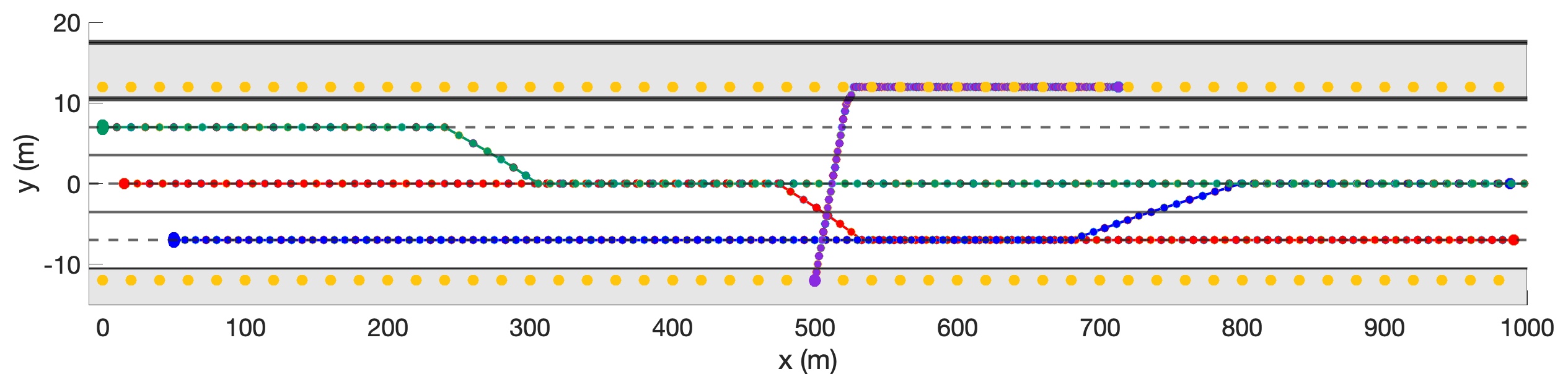}
    \caption{Schematic for the ground truth trajectory of the driving example. The ego vehicle (red) navigates and runs dynamic EKF SLAM along a kilometer-long stretch of highway, from left to right, alongside two other vehicles (green, blue) and a jaywalking pedestrian (purple). Static landmarks (yellow) are scattered throughout the scene. Initial feature estimates of each moving object are plotted, but are not clearly visible due to the schematic scale.}
    \label{fig:Highway_Schematic}
\end{figure*}

\begin{table*}
    \centering
    \vspace{5mm}
    \begin{tabular}{||c| c| c c c c c c c  c||} 
         \hline
         Noise & Data & Ego & Static & Agent 1 & Agent 2 & Agent 3 & Agent 1 & Agent 2 & Agent 3 \\ 
         Level & & Poses & Features & Features & Features & Features & Poses & Poses & Poses \\ 
         \hline
         \hline
         $\Sigma_{w,1}, \Sigma_{v,1}$ & $x$ (m) & 0.008 & 0.014 & 0.014 & 0.014 & 0.014 & 0.001 & 0.001 & 0.001 \\ 
         & $y$ (m) & 0.032 & 0.028 & 0.003 & 0.001 & 0.025 & 0.028 & 0.035 & 0.006 \\
         & $\theta$ (rad) & 0.000 & N/A & N/A & N/A & N/A & 0.001 & 0.001 & 0.006 \\
         \hline
         $\Sigma_{w,1}, \Sigma_{v,2}$ & $x$ (m) & 0.035 & 0.062 & 0.062 & 0.062 & 0.063 & 0.002 & 0.002 & 0.002 \\ 
         & $y$ (m) & 0.061 & 0.059 & 0.007 & 0.003 & 0.052 & 0.054 & 0.066 & 0.012 \\
         & $\theta$ (rad) & 0.000 & N/A & N/A & N/A & N/A & 0.004 & 0.004 & 0.015 \\
         \hline
         $\Sigma_{w,1}, \Sigma_{v,3}$ & $x$ (m) & 0.089 & 0.155 & 0.157 & 0.157 & 0.156 & 0.005 & 0.004 & 0.006 \\ 
         & $y$ (m) & 0.152 & 0.157 & 0.168 & 0.009 & 0.138 & 0.137 & 0.166 & 0.032 \\
         & $\theta$ (rad) & 0.000 & N/A & N/A & N/A & N/A & 0.011 & 0.011 & 0.054 \\
         \hline
         $\Sigma_{w,2}, \Sigma_{v,1}$ & $x$ (m) & 0.008 & 0.012 & 0.012 & 0.012 & 0.012 & 0.002 & 0.002 & 0.003 \\ 
         & $y$ (m) & 0.072 & 0.053 & 0.004 & 0.004 & 0.046 & 0.065 & 0.080 & 0.014 \\
         & $\theta$ (rad) & 0.000 & N/A & N/A & N/A & N/A & 0.001 & 0.001 & 0.005 \\
         \hline
         $\Sigma_{w,2}, \Sigma_{v,2}$ & $x$ (m) & 0.028 & 0.048 & 0.049 & 0.048 & 0.050 & 0.003 & 0.003 & 0.004 \\ 
         & $y$ (m) & 0.088 & 0.0074 & 0.007 & 0.005 & 0.065 & 0.078 & 0.004 & 0.018 \\
         & $\theta$ (rad) & 0.000 & N/A & N/A & N/A & N/A & 0.004 & 0.004 & 0.016 \\
         \hline
         $\Sigma_{w,2}, \Sigma_{v,3}$ & $x$ (m) & 0.103 & 0.181 & 0.182 & 0.182 & 0.182 & 0.005 & 0.005 & 0.007 \\ 
         & $y$ (m) & 0.200 & 0.196 & 0.020 & 0.010 & 0.173 & 0.180 & 0.217 & 0.041 \\
         & $\theta$ (rad) & 0.000 & N/A & N/A & N/A & N/A & 0.012 & 0.012 & 0.053 \\
         \hline
         $\Sigma_{w,3}, \Sigma_{v,1}$ & $x$ (m) & 0.017 & 0.017 & 0.019 & 0.016 & 0.021 & 0.009 & 0.008 & 0.012 \\ 
         & $y$ (m) & 0.287 & 0.232 & 0.016 & 0.012 & 0.202 & 0.259 & 0.317 & 0.058 \\
         & $\theta$ (rad) & 0.001 & N/A & N/A & N/A & N/A & 0.001 & 0.001 & 0.005 \\
         \hline
         $\Sigma_{w,3}, \Sigma_{v,2}$ & $x$ (m) & 0.029 & 0.043 & 0.044 & 0.044 & 0.045 & 0.010 & 0.009 & 0.013 \\ 
         & $y$ (m) & 0.226 & 0.164 & 0.015 & 0.009 & 0.144 & 0.200 & 0.246 & 0.044 \\
         & $\theta$ (rad) & 0.001 & N/A & N/A & N/A & N/A & 0.004 & 0.004 & 0.016 \\
         \hline
         $\Sigma_{w,3}, \Sigma_{v,3}$ & $x$ (m) & 0.093 & 0.161 & 0.161 & 0.010 & 0.014 & 0.010 & 0.010 & 0.014 \\ 
         & $y$ (m) & 0.312 & 0.264 & 0.027 & 0.014 & 0.233 & 0.277 & 0.340 & 0.061 \\
         & $\theta$ (rad) & 0.001 & N/A & N/A & N/A & N/A & 0.012 & 0.013 & 0.053 \\
         \hline
        \end{tabular}
    \caption{Root-mean-squared translation ($x, y$) and rotation ($\theta$) error on our simulated driving dataset. Noise level settings correspond to different choices of $\Sigma_w$ and $\Sigma_v$ (with $\Sigma_0 = \Sigma_w$), 
    as defined in \ref{Eqn: Sigma w, v, 1}, \ref{Eqn: Sigma w, v, 2}, \ref{Eqn: Sigma w, v, 3}. Root-mean-squared errors are averaged over 25 experiments for each noise setting.
    }
    \label{tab:RMS_Error}
\end{table*}



\begin{theorem} \label{Thm: Dynamic EKF, Dynamic Object Pose Augmentation}
The dynamic object pose augmentation step of standard-formulation dynamic EKF SLAM (Alg. \ref{Alg: Dynamic EKF, Moving Object Pose Augmentation, Standard}) is equivalent to applying a Gauss-Newton step to $c_{\dEKF,t,3}: \R^{\dmu + \no \dx} \ra \R$, with:
\begin{align*}
    &c_{\dEKF,t,3}(\tilde{x}_t, \fpose_{t, 1}, \cdots, \fpose_{t, \no}) \\
    := \hspace{0.5mm} &\Vert \tilde{x}_t - \mu_t \Vert_{\Sigma_t^{-1}}^2 + \sum_{\alpha=1}^{\no} \sum_{k=1}^{\nof(\alpha)} \Vert \fd_{t, \alpha, k} - \go (\fpose_{t, \alpha}, \fd_{0, \alpha, k}) \Vert_{\Sigma_\fpose^{-1}}^2.
\end{align*}
when $\Sigma_{\fpose}$ is a diagonal matrix.
\end{theorem}

\begin{proof}
Please see Appendix \ref{subsec: A1, Proofs of Main Theorems} in the extended version of this paper \cite{chiu2022SLAMBackends}.
\end{proof}




\begin{theorem} \label{Thm: Dynamic EKF, Smoothing Update}
The smoothing update step of standard-formulation dynamic EKF SLAM (Alg. \ref{Alg: Dynamic EKF, Smoothing Update, Standard}) is equivalent to applying a Gauss-Newton step to $c_{\dEKF,t,7}: \R^{\dmu} \ra \R$, with:
\begin{align*}
    &c_{\dEKF,t,7}(\tilde{x}_t) \\
    := \hspace{0.5mm} &\Vert \tilde{x}_t - \mu_t \Vert_{\Sigma_t^{-1}}^2 + \sum_{\alpha=1}^{\no} \Vert s(\fpose_{t-2, \alpha}, \fpose_{t-1, \alpha}, \fpose_{t, \alpha}) \Vert_{\Sigma_s^{-1}}^2.
\end{align*}
\end{theorem}

\begin{proof}
Please see Appendix \ref{subsec: A1, Proofs of Main Theorems} in the extended version of this paper \cite{chiu2022SLAMBackends}.
\end{proof}




\section{Experiments}
\label{sec: Experiments}


To illustrate the estimation accuracy and mapping precision of the dynamic EKF algorithm presented above, we constructed a simulated driving scenario (Figure \ref{fig:Highway_Schematic}). In the scenario, the ego vehicle navigates alongside two other vehicles (Agents 1, 2) and a pedestrian (Agent 3) on a highway with three lanes, while simultaneously tracking the positions of non-ego vehicles and fixed landmarks in its surroundings. As time progresses, the vehicles change lanes and adjust their velocities. Object motion is sampled every 0.5 s for 60 s to form a ground truth dataset. 

To test our dynamic EKF algorithm, we performed Monte Carlo experiments on the simulated driving setting described above. For each combination of the three odometry and image measurement noise covariance levels given below, we simulated the ground truth trajectory 25 times, each with independently generated errors:
\begin{alignat}{2} \label{Eqn: Sigma w, v, 1}
    \Sigma_{w,1} &:= \begin{bmatrix}
    10^{-6} & 0 & 0 \\
    0 & 10^{-6} & 0 \\
    0 & 0 & 10^{-8}
    \end{bmatrix}, 
    \Sigma_{v,1} &&:= 10^{-6} \cdot I_{2 \times 2}, \\ \label{Eqn: Sigma w, v, 2}
    \Sigma_{w,2} &:= \begin{bmatrix}
    10^{-5} & 0 & 0 \\
    0 & 10^{-5} & 0 \\
    0 & 0 & 10^{-7}
    \end{bmatrix}, 
    \Sigma_{v,2} &&:= 10^{-5} \cdot I_{2 \times 2}, \\ \label{Eqn: Sigma w, v, 3}
    \Sigma_{w,3} &:= \begin{bmatrix}
    10^{-4} & 0 & 0 \\
    0 & 10^{-4} & 0 \\
    0 & 0 & 10^{-6}
    \end{bmatrix}, 
    \Sigma_{v,3} &&:= 10^{-4} \cdot I_{2 \times 2}, 
\end{alignat}
Here, $I_{2 \times 2}$ denotes the $2 \times 2$ identity matrix, and each entry of the above matrices has unit $m^2$ (meters squared). We then applied dynamic EKF SLAM to recover the ground truth trajectory, and computed the resulting root-mean-squared error for each noise level (Table \ref{tab:RMS_Error}). By dropping past poses of all moving objects, each simulation can be run on a standard, single-threaded laptop in under 37 ms. We observed that our estimation accuracy decreases gracefully as the noise level increased. For more details regarding the simulation setup and results, please see Appendix \ref{subsec: A3, Experiment Details}.




\begin{algorithm} \label{Alg: Dynamic EKF, Feature Augmentation, Standard}
    {
    \small
    \SetAlgoLined
    \KwData{Current EKF state $\tilde x_t \in \R^{\dmu}$, with mean $\mu_t$, and covariance $\Sigma_t$; number of static features $\nf \in \N$; total number of features on moving objects $\Nf \in \N$; 
    measurements of new features $z_{t,k} \in \R^{\dz}, k \in [\nf+\Nf+1: \nf+\Nf+\nf'+\Nf']$; measurement map $h: \R^{\dx} \times \R^{\df} \ra \R^{\dz}$; inverse measurement map $\ell: \R^{\dx} \times \R^{\dz} \ra \R^{\df}$, with $z_{t,k} = h \big( \xrt, \ell(\xrt, z_{t,k}) \big) \hspace{0.5mm} \forall \hspace{0.5mm} \xrt \in \R^{\dx}$.}
    
    \KwResult{Updated number of static features $\nf$; updated total number of features on moving objects $\Nf$, updated EKF state dimension $\dmu$, updated EKF state mean $\mu_t \in \R^{\dmu}$, covariance $\Sigma_t \in \R^{\dmu \times \dmu}$}
    
    $(\mu_{t,x}, \mu_{t,e}) \gets \mu_t$, with $\mu_{t,x} \in \R^{\dx}$.
     
    
    $\tilde \ell(\mu_{t,x}, z_{t,\nf+\Nf+1}, \cdots, z_{t,\nf+\Nf+\nf'+\Nf'}) \gets \big( \ell(\mu_{t,x}, z_{t,\nf+\Nf+1}), \cdots, \ell(\mu_{t,x}, z_{t,\nf+\Nf+\nf'+\Nf'}) \big) \in \R^{(\nf'+\Nf') \df}$, with $\tilde \ell: \R^{\dx + \Nf' \dz} \ra \R^{\Nf' \df}$
    
    $\mu_t \gets \big(\mu_t,  \tilde\ell(\mu_{t,x}, z_{t,\nf'+\Nf'+1}, \cdots, z_{t,\nf+\Nf+\nf'+\Nf'}) \big) \in \R^{\dmu + (\nf'+\Nf') \df}$
    
    $\begin{bmatrix}
    \Sigma_{t,xx} & \Sigma_{t,xe} \\
    \Sigma_{t,ex} & \Sigma_{t,ee} 
    \end{bmatrix} \gets \Sigma_t$, with $\Sigma_{t,xx} \in \R^{\dx \times \dx}$
    
    $L_x \gets \frac{\partial \tilde \ell}{\partial x} \big|_{\mu_t} \in \R^{(\nf'+\Nf') \df \times \dx}$
    
    $L_z \gets \frac{\partial \tilde \ell}{\partial z} \big|_{\mu_t} \in \R^{(\nf'+\Nf')\df \times (\nf'+\Nf')\dz}$
    
    $\tilde{\Sigma}_v \gets \text{diag}\{\Sigma_v, \cdots, \Sigma_v\} \in \R^{(\nf'+\Nf')\dz \times (\nf'+\Nf')\dz}$
    
    $\Sigma_t \gets \begin{bmatrix}
    \Sigma_{t,xx} & \Sigma_{t,xe} & \Sigma_{t,xx} L_x^\top \\
    \Sigma_{t,ex} & \Sigma_{t,ee} & \Sigma_{t,ex} L_x^\top \\
    L_x \Sigma_{t,xx} & L_x \Sigma_{t,xe} & L_x \Sigma_{t,xx} L_x^\top + L_z \tilde{\Sigma}_v L_z^\top
    \end{bmatrix} \in \R^{(\dmu + (\nf'+\Nf') \df) \times (\dmu + (\nf'+\Nf') \df)} $
    
    $\tilde x_t \gets (\tilde x_t, f_{t, \Nf+1}, \cdots, f_{t, \Nf+\Nf'}) \in \R^{\dmu + (\nf'+\Nf') \df}$
    
    $\nf \gets \nf + \nf'$
    
    $\Nf \gets \Nf + \Nf'$
    
    $\dmu \gets \dmu + (\nf' + \Nf') \df$
    
    \label{Eqn: Dynamic EKF, Feature Augmentation, Variable Reordering}
    Reorder variables in $\tilde x_t$ to restore the variable ordering of \eqref{Eqn: Dynamic EKF SLAM, Full State}.
    
    \Return{$\Nf, \dmu, \mu_t, \Sigma_t$.}
     
     \caption{Dynamic EKF, Feature Augmentation Sub-block.}
     }
\end{algorithm}


\begin{algorithm} \label{Alg: Dynamic EKF, Moving Object Pose Augmentation, Standard}
    {
    \small
    \SetAlgoLined
    \KwData{Current EKF state $\tilde x_t \in \R^{\dmu}$ mean $\mu_t$ and covariance $\Sigma_t$; number of dynamic objects $\no$ with features detected at times $0$ and $t$; moving object dynamics map $\go: \R^{\dx} \times \R^{\df} \ra \R^{\df}$ for each object indexed $\alpha \in [\no]$; inverse moving object dynamics map $\invg^\alpha: \R^{\nof(\alpha) \df} \times \R^{\nof(\alpha) \df} \ra \R^{\dx}$, satisfying $\xi = \invg^\alpha(f, \go(\xi, f)) $ for each $f, f' \in \R^{\nof(\alpha) \df}$; moving object dynamics noise covariance $\tilde \Sigma_\fpose \in \R^{\df \times \df}$; number of features $\nof(\alpha) \in \N$ detected on each moving object with index $\alpha \in [\no]$.}
    
    \KwResult{Updated EKF state dimension $\dmu$, updated EKF state mean $\mu_t \in \R^{\dmu}$, covariance $\Sigma_t \in \R^{\dmu \times \dmu}$.}
    
    
    $\Nf \gets \sum_{\alpha=1}^{\no} \nof(\alpha)$
    
    $\fd_{\tau, \alpha} \gets (\fd_{\tau, \alpha, 1}, \cdots, \fd_{\tau, \alpha, \nof(\alpha)}) \in \R^{\nof(\alpha) \cdot \df}$, $\forall \hspace{0.5mm} \alpha \in [\no], \tau \in \{0, t\}$
    
    $\fd_{\tau} \gets (\fd_{\tau, 1}, \cdots, \fd_{\tau, \no}) \in \R^{\Nf \df}, \forall \hspace{0.5mm} \tau \in \{0, t\}$
    
    $\xi_t \gets (\xi_{t, 1}, \cdots, \xi_{t, \no}) \in \R^{\no \dx}$.
    
    \label{Eqn: Dynamic EKF, Moving Object Pose Augmentation, Variable Reordering, 1}
    Reorder variables in $\tilde x_t$ so that $\fd_0$, $\fd_t$ appear last.
    
    $\begin{bmatrix}
        \Sigma_{t, ee} & \Sigma_{t, ef} & \Sigma_{t, ef'} \\
        \Sigma_{t, fe} & \Sigma_{t, ff} & \Sigma_{t, ff'}  \\
        \Sigma_{t, f' e} & \Sigma_{t, f' f} & \Sigma_{t, f' f'}
    \end{bmatrix} \gets \Sigma_t$, with $\Sigma_{t, ff}, \Sigma_{t, f'f'} \in \R^{\Nf \df \times \Nf \df}$
    
    $\tilde \invg( \fd_{0}, \fd_{t}) \gets \big( \invg^1(\fd_{0, 1}, \fd_{t, 1}), \cdots, \invg^\no(\fd_{0, \no}, \fd_{t, \no}) \big) \in \R^{\no \dx}$, with $\tilde \invg: \R^{\Nf \df} \times \R^{\Nf \df} \ra \R^{\no \dx}$
    
    $\tilde \go(\xi_t, \fd_0) \gets \big( \go(\xi_{t, 1}, \fd_{0, 1}, \cdots, \xi_{t, \no}, \fd_{0, \no}) \big) \in \R^{\Nf \df}$
    
    $\tilde \Gxio \gets \frac{\partial \tilde \go}{\partial \xi}(\xi_t, \fd_0)$
    
    $\Tilde \Invg_1 \gets \frac{\partial \tilde \invg}{\partial \fd_{0, \no}} \Big|_{\fd_{0, \no}} \in \R^{\no \dx \times \Nf \df}$
    
    $\Tilde \Invg_2 \gets \frac{\partial \tilde \invg}{\partial \fd_{t, \no}} \Big|_{\fd_{t, \no}} \in \R^{\no \dx \times \Nf \df}$
    
    $\mu_t \gets \big( \mu_t, \tilde \invg( \fd_{0}, \fd_{t} ) \big) \in \R^{\dmu + \no \dx}$
    
    $\Sigma_{t, \fpose e} \gets \tilde \Invg_1 \Sigma_{t, fe} + \tilde \Invg_2 \Sigma_{t, f' e} \in \R^{\no \dx \times (\dmu - 2 \Nf \df)}$
    
    $\Sigma_{t, \fpose f} \gets \tilde \Invg_1 \Sigma_{t, ff} + \tilde \Invg_2 \Sigma_{t, f' f} \in \R^{\no \dx \times \Nf \df}$
    
    $\Sigma_{t, \fpose f'} \gets \tilde \Invg_1 \Sigma_{t, ff'} + \tilde \Invg_2 \Sigma_{t, f' f'} \in \R^{\no \dx \times \Nf \df}$
    
    $\Sigma_{t, \fpose \fpose} \gets \tilde \Invg_1 \Sigma_{t, ff} \tilde \Invg_1^\top + \tilde \Invg_2 \Sigma_{t, f' f} \tilde \Invg_1^\top + \tilde \Invg_1 \Sigma_{t, f f'} \tilde \Invg_2^\top + \tilde \Invg_2 (\Sigma_{t, f' f'} + \tilde \Sigma_\fpose) \tilde \Invg_2^\top 
    \in \R^{\no \dx \times \no \dx}$
    
    $\Sigma_t \gets \begin{bmatrix}
        \Sigma_{t, ee} & \Sigma_{t, ef} & \Sigma_{t, ef'} & \Sigma_{t, \fpose e}^\top \\
        \Sigma_{t, fe} & \Sigma_{t, ff} & \Sigma_{t, ff'} & \Sigma_{t, \fpose f}^\top \\
        \Sigma_{t, f' e} & \Sigma_{t, f' f} & \Sigma_{t, f' f'} & \Sigma_{t, \fpose f'}^\top \\
        \Sigma_{t, \fpose e} & \Sigma_{t, \fpose f} & \Sigma_{t, \fpose f'} & \Sigma_{t, \fpose \fpose}
    \end{bmatrix} \in \R^{(\dmu + \no \dx) \times (\dmu + \no \dx)}$
    
    $\dmu \gets \dmu + \no \dx$
    
    \label{Eqn: Dynamic EKF, Moving Object Pose Augmentation, Variable Reordering, 2}
    Reorder variables in $\tilde x_t$ to restore the variable ordering of \eqref{Eqn: Dynamic EKF SLAM, Full State}.
    
    \Return{$\dmu$, $\mu_t$, $\Sigma_t$.}
     
     \caption{Dynamic EKF, Moving Object Pose Augmentation Sub-block.}
     }
\end{algorithm}


\begin{algorithm} \label{Alg: Dynamic EKF, Static Feature Update, Standard}
    {
    \small
    \SetAlgoLined
    
    \KwData{Prior $\mathcal N(\mu_0, \Sigma_0)$ on $\xr_0 \in \R^{d_x}$, noise covariances $\Sigma_w \in \R^{\dx \times \dx}$, $\Sigma_v \in \R^{\dz \times \dz}$, $\Sigma_s \in \R^{\dx \times \dx}$, dynamics map $g: \R^{\dx } \ra \R^{\dx}$, measurement map $h: \R^{\dx} \times \R^{\df} \ra \R^{dz}$, inverse measurement map $\ell: \R^{\dx} \times \R^{\dz} \ra \R^{df}$; time horizon $T \in \N$; number of features $\nf \in \N$; number of moving objects $\no \in \N$.}
    
    \KwResult{Updated EKF state mean $\mu_t \in \R^{\dmu}$, covariance $\Sigma_t \in \R^{\dmu \times \dmu}$.}
    
    $\fs \gets (\fs_{1}, \cdots, \fs_{\nf}) \in \R^{\nf \df}$
    
    $\Tilde{h}(\xrt, \fs) \gets \big( \hs(\xrt, \fs_{1}), \cdots, \hs(\xrt, \fs_{\Nf}) \big) \in \R^{\nf \dz}$
        
        $H_t \gets \frac{\partial \tilde{h}}{\partial (\xrt, \fs)} \Big|_{\mu_t} \in \R^{\nf dz \times (\dx + \nf \df)}$
        
        $\Tilde{\Sigma}_v \gets \text{diag}\{\Sigma_v, \cdots, \Sigma_v \} \in \R^{\nf \dz \times \nf \dz}$
        
        \label{Eqn: Dynamic EKF, Mean Update}
        $\overline{\mu_t} \gets \mu_t + \Sigma_t \begin{bmatrix}
            H_t^\top \\ O
        \end{bmatrix} \left( \begin{bmatrix}
            H_t & O
        \end{bmatrix} \Sigma_t \begin{bmatrix}
            H_t^\top \\ O
        \end{bmatrix} + \Tilde{\Sigma}_v \right)^{-1} \big(z_{t,1:\nf} - \tilde{h}(\mu_t, \fs) \big) \in \R^{\dmu}$
        
        \label{Eqn: Dynamic EKF, Cov Update}
        $\overline{\Sigma}_t \gets \Sigma_t - \Sigma_t \begin{bmatrix}
            H_t^\top \\ O
        \end{bmatrix} \left(\begin{bmatrix}
            H_t & O
        \end{bmatrix} \Sigma_t \begin{bmatrix}
            H_t^\top \\ O
        \end{bmatrix} + \Tilde{\Sigma}_v \right)^{-1} \begin{bmatrix}
            H_t & O
        \end{bmatrix} \Sigma_t \in \R^{\dmu \times \dmu}$
    
    \Return{$\mu_t, \Sigma_t$.}
     
     \caption{Dynamic EKF, Static Feature Update Sub-block.}
     }
\end{algorithm}


\begin{algorithm} \label{Alg: Dynamic EKF, Smoothing Update, Standard}
    {
    \small
    \SetAlgoLined
    \KwData{Current EKF state $\tilde x_t \in \R^{\dmu}$, with mean $\mu_t$ and covariance $\Sigma_t$, smoothing covariance $\Sigma_s \in \R^{\dx \times \dx}$, number of dynamic objects $\alpha \in [n_o]$ with poses $\xi_{\tau, \alpha}$ tracked at times $\tau \in \{ t-2, t-1, t\}$.}
    
    \KwResult{Updated EKF state mean $\mu_t \in \R^{\dmu}$, covariance $\Sigma_t \in \R^{\dmu \times \dmu}$.}
    
    $\fpose_{\tau} \gets (\fpose_{\tau, 1}, \cdots, \fpose_{\tau}, \no) \in \R^{\no \dx}, \forall \hspace{0.5mm} \tau \in \{t-2, t-1, t\}$
    
    \label{Eqn: Dynamic EKF, Smoothing Update, Variable Reordering, 1}
    Reorder variables in $\tilde x_t$ such that $(\fpose_{t-2}, \fpose_{t-1}, \fpose_t) \in \R^{3 \no \dx}$ appears last.
    
    $\tilde \Sigma_s \gets \text{diag}\{\Sigma_s, \cdots, \Sigma_s\} \in \R^{\no \dx \times \no \dx}$
    
    $\tilde s(\xi_{t-2}, \xi_{t-1}, \xi_t) \gets \big( s(\xi_{t-2, 1}, \xi_{t-1, 1}, \xi_{t, 1}), \cdots, s(\xi_{t-2, \no}, \xi_{t-1, \no}, \xi_{t, \no}) \big) \in \R^{\no \dx}$
    
    $\tilde S_\tau \gets \frac{\partial \tilde s}{\partial \xi_\tau}$, for each $\tau \in \{t-2, t-1, t\}$.
    
      
    $\mu_t \gets \mu_t - \Sigma_t \begin{bmatrix}
        O \\ \tilde S_{t-2}^\top \\ \tilde S_{t-1}^\top \\ \tilde S_t^\top
    \end{bmatrix} \left(\Tilde{\Sigma}_s + \begin{bmatrix}
        O & \tilde S_{t-2} & \tilde S_{t-1} & \tilde S_t
    \end{bmatrix} \Sigma_t^{-1}    \begin{bmatrix}
        O \\ \tilde S_{t-2}^\top \\ \tilde S_{t-1}^\top \\ \tilde S_t^\top
    \end{bmatrix} \right)^{-1} \cdot s(\fpose_{t-2}, \fpose_{t-1}, \fpose_t) \in \R^{\dmu}$
    
    $\Sigma_t \gets \Sigma_t - \Sigma_t \begin{bmatrix}
        O \\ \tilde S_{t-2}^\top \\ \tilde S_{t-1}^\top \\ \tilde S_t^\top
    \end{bmatrix} \left(\Tilde{\Sigma}_s + \begin{bmatrix}
        O & \tilde S_{t-2} & \tilde S_{t-1} & \tilde S_t
    \end{bmatrix} \Sigma_t^{-1}    \begin{bmatrix}
        O \\ \tilde S_{t-2}^\top \\ \tilde S_{t-1}^\top \\ \tilde S_t^\top
    \end{bmatrix} \right)^{-1} \newline
    \cdot \begin{bmatrix}
        O & \tilde S_{t-2} & \tilde S_{t-1} & \tilde S_t
    \end{bmatrix} \Sigma_t \in \R^{\dmu \times \dmu}$
    
    \label{Eqn: Dynamic EKF, Smoothing Update, Variable Reordering, 2}
    Reorder variables in $\tilde x_t$ to restore the variable ordering of \eqref{Eqn: Dynamic EKF SLAM, Full State}.
    
    \Return{$\mu_t$, $\Sigma_t$.}
     
     \caption{Dynamic EKF, Smoothing Update Sub-block.}
     }
\end{algorithm}


\begin{algorithm} \label{Alg: Dynamic EKF, State Propagation, Standard}
    {
    \small
    \SetAlgoLined
    \KwData{Current EKF state $\tilde x_t \in \R^{\dmu}$, with mean $\overline{\mu_t}$ and covariance $\overline{\Sigma_t}$, (discrete-time) dynamics map $g: \R^{\dx} \ra \R^{\dx}$.}
    
    \KwResult{Propagated EKF state mean $\mu_{t+1} \in \R^{\dmu}$ and covariance $\Sigma_{t+1} \in \R^{\dmu \times \dmu}$}
    
    $(\overline{\mu}_{t,x}, \overline{\mu}_{t,e}) \gets \overline{\mu_t}$, with $\overline{\mu}_{t,x} \in \R^{\dx} := $ ego robot pose mean.
            
    $\begin{bmatrix}
        \overline{\Sigma}_{t,xx} & \overline{\Sigma}_{t,xe} \\
        \overline{\Sigma}_{t,ex} &
        \overline{\Sigma}_{t,ee}
    \end{bmatrix} \gets \overline{\Sigma}_t$, with $\overline{\Sigma}_{t,xx} \in \R^{\dx \times \dx} := $ ego robot pose covariance.
    

    $G_t \gets \frac{\partial g}{\partial x} \Big|_{\overline{\mu}_{t,x}} \in \R^{\dx \times \dx}$.
    
    $\mu_{t+1} \gets \big( g(\overline{\mu}_{t,x}), \overline{\mu}_{t,e} \big) \in \R^{\dmu}.$  \label{Eqn: Dynamic EKF, Mean Propagation}
    
    $\Sigma_{t+1} \gets \begin{bmatrix}
        G_t \overline{\Sigma}_{t,xx} G_t^\top + \Sigma_w & G_t \overline{\Sigma}_{t,xe} \\
        \overline{\Sigma}_{t,ex} G_t^\top & \overline{\Sigma}_{t,ee}
    \end{bmatrix} \in \R^{\dmu \times \dmu}.$ \label{Eqn: Dynamic EKF, Cov Propagation}
    
    \Return{$\mu_{t+1}, \Sigma_{t+1}$.}
     
     \caption{Dynamic EKF, State Propagation Sub-block.}
     }
\end{algorithm}

\section{Conclusion and Future Work}
\label{sec: Conclusion and Future Work}

In this tutorial, we extended the unifying optimization-based SLAM backend framework in \cite{Saxena2022SLAMNonlinearOptimization} to environments with moving objects. We use this framework to describe the back-ends of recently proposed dynamic SLAM algorithms \cite{Yang2019CubeSLAM, Huang2019ClusterSLAM, zhang2020VDOSLAM, Bescos2021DynaSLAMII}. To establish a rapprochement with filtering-based SLAM methods, we apply an aggressive marginalization scheme in our framework to derive the dynamic EKF SLAM algorithm, which we prove to be mathematically identical to the straightforward extension of the conventional EKF-SLAM algorithm to environments with moving objects. Simulation results indicate that dynamic EKF-SLAM performs well in pose estimation, as well as static and dynamic feature tracking.

The formulation presented in this tutorial can be refined in several ways. First, we are eager to deploy our framework on real-world data, and explore the tradeoffs inherent in different design choices. Second, many robotics applications require topological and/or semantic maps of dynamic scenes, in addition to purely metric information. Thus, dynamic object features should be explicitly encoded in our framework as lower-dimensional semantic representations, e.g., bounding boxes, as is done in existing methods \cite{zhang2020VDOSLAM,  Bescos2021DynaSLAMII, Yang2019CubeSLAM}. Third, robust formulations of our framework must allow for the implementation of multi-hypthesis SLAM backends, to reliably safeguard against ambiguous data associations or high outlier densities \cite{Doherty2020ProbabilisticDataAssociationviaMixtureModelsforRobustSemanticSLAM, HsiaoKaess2019MHiSAM2}.
Finally, guidelines for selecting appropriate modeling choices within our dynamic SLAM framework depend heavily on downstream tasks, such as autonomous navigation in the presence of multiple agents \cite{fridovich2019efficient, laine2021computation}. It is of interest to design complete autonomy stacks that fully harness the flexibility of our dynamic SLAM framework for estimation, prediction, and planning in challenging robotics tasks.



\printbibliography

@article{Leutenegger2015KeyframebasedVO,
  title = {{Keyframe-based Visual-Inertial Odometry using Nonlinear Optimization}},
  author={S. Leutenegger and S. Lynen and M. Bosse and R. Siegwart and P. Furgale},
  journal={IJRR},
  year={2015},
  volume={34},
  pages={314 - 334}
}

@book{Thrun2005ProbabilisticRobotics,
author = {Sebastian Thrun and Wolfram Burgard and Dieter Fox},
title = {{Probabilistic Robotics}},
year = {2005},
% isbn = {0262201623},
publisher = {The MIT Press}
}

@article{Sola2014SLAMWithEKF,
    author = {Joan Sol\`{a}},
    title = {{Simultaneous Localization and Mapping with the Extended Kalman Filter}},
    journal = {arXiv},
    eprint={1803.11288},
    year = {2014}
}

@article{Cadena2016PastPresentandFuture,
  author={C. {Cadena} and L. {Carlone} and H. {Carrillo} and Y. {Latif} and D. {Scaramuzza} and J. {Neira} and I. {Reid} and J. J. {Leonard}},
  journal={IEEE T-RO}, 
  title = {{Past, Present, and Future of Simultaneous Localization and Mapping: Toward the Robust-Perception Age}}, 
  year={2016},
  volume={32},
  number={6},
  pages={1309-1332},
}

@article{laine2021computation,
  title = {{The Computation of Approximate Generalized Feedback Nash Equilibria}},
  author={Forrest Laine and David Fridovich-Keil and Chih-Yuan Chiu and Claire Tomlin},
  journal={arXiv},
  year={2021}
}

@article{fridovich2019efficient,
  author={Fridovich-Keil, David and Ratner, Ellis and Peters, Lasse and Dragan, Anca D. and Tomlin, Claire J.},
  booktitle={2020 IEEE International Conference on Robotics and Automation (ICRA)}, 
  title = {{Efficient Iterative Linear-Quadratic Approximations for Nonlinear Multi-Player General-Sum Differential Games}},
  year={2020},
  volume={},
  number={},
  pages={1475-1481},
  doi={10.1109/ICRA40945.2020.9197129}}

@article{zhang2020VDOSLAM,
      title = {{VDO-SLAM: A Visual Dynamic Object-aware SLAM System}}, 
      author={Jun Zhang and Mina Henein and Robert Mahony and Viorela Ila},
      year={2020},
      booktitle={arXiv},
      eprint={2005.11052},
      archivePrefix={arXiv},
      primaryClass={cs.RO}
}

@article{LeonardDurrantWhyte1991SimultaneousMapBuilding,
  author={Leonard, J.J. and Durrant-Whyte, H.F.},
  booktitle={IEEE IROS}, 
  title={{Simultaneous Map Building and Localization for an Autonomous Mobile Robot}}, 
  year={1991},
  volume={3},
  number={},
  pages={1442-7},
%   doi={10.1109/IROS.1991.174711}
}

@article{Saxena2022SLAMNonlinearOptimization,
  author={Saxena, Amay and Chiu, Chih-Yuan and Shrivastava, Ritika and Menke, Joseph and Sastry, Shankar},
  journal={IEEE Robotics and Automation Letters}, 
  title={{Simultaneous Localization and Mapping: Through the Lens of Nonlinear Optimization}}, 
  year={2022},
  volume={7},
  number={3},
  pages={7148-7155},
  doi={10.1109/LRA.2022.3181409}}

@book{Dontchev2005ImplicitFunctionsAndSolutionMappings,
author = {Asen Dontchev and Rockafellar Tyrrell},
title = {{Implicit Functions and Solution Mappings: A View from Variational Analysis}},
year = {2009},
isbn = {978-0-387-87820-1},
publisher = {Springer Science Business Media, LLC}
}

@article{bescos2018dynaslam,
  author = {Bescos, Berta and F{\'a}cil, Jos{\'e} M. and Civera, Javier and Neira, Jos{\'e}},
  title = {{DynaSLAM: Tracking, Mapping and Inpainting in Dynamic Scenes}},
  journal = {Robotics and Automation Letters RA-L},
  year = {2018}
}

@article{Bescos2021DynaSLAMII,
  title={DynaSLAM II: Tightly-Coupled Multi-Object Tracking and SLAM},
  author={Berta Besc{\'o}s and Carlos Campos and Juan D. Tard{\'o}s and Jos{\'e} Neira},
  journal={IEEE Robotics and Automation Letters},
  year={2021},
  volume={6},
  pages={5191-5198}
}

@article{Yang2019CubeSLAM,
  author={Yang, Shichao and Scherer, Sebastian},
  journal={IEEE Transactions on Robotics}, 
  title={{CubeSLAM: Monocular 3-D Object SLAM}}, 
  year={2019},
  volume={35},
  number={4},
  pages={925-938},
  doi={10.1109/TRO.2019.2909168}}

@article{Wang2007SLAMMOT,
author = {Chieh-Chih Wang and Charles Thorpe and Sebastian Thrun and Martial Hebert and Hugh Durrant-Whyte},
title ={{Simultaneous Localization, Mapping and Moving Object Tracking}},
journal = {The International Journal of Robotics Research},
volume = {26},
number = {9},
pages = {889-916},
year = {2007},
doi = {10.1177/0278364907081229},

URL = { 
        https://doi.org/10.1177/0278364907081229
    
},
eprint = { 
        https://doi.org/10.1177/0278364907081229
    
}
,
 
}

@article{Doherty2020ProbabilisticDataAssociationviaMixtureModelsforRobustSemanticSLAM,
author = {Doherty, Kevin and Baxter, David and Schneeweiss, Edward and Leonard, John},
year = {2020},
month = {05},
pages = {1098-1104},
title = {{Probabilistic Data Association via Mixture Models for Robust Semantic SLAM}},
doi = {10.1109/ICRA40945.2020.9197382}
}

@article{HsiaoKaess2019MHiSAM2,
  title={{MH-iSAM2: Multi-hypothesis iSAM using Bayes Tree and Hypo-tree}},
  author={Ming Hsiao and Michael Kaess},
  journal={2019 International Conference on Robotics and Automation (ICRA)},
  year={2019},
  pages={1274-1280}
}

@article{Huang2019ClusterSLAM,
author = {Huang, Jiahui and Yang, Sheng and Zhao, Zishuo and Lai, Yu-Kun and Hu, Shi-Min},
title = {{ClusterSLAM: A SLAM Backend for Simultaneous Rigid Body Clustering and Motion Estimation}},
booktitle = {Proceedings of the IEEE/CVF International Conference on Computer Vision (ICCV)},
year = {2019}
}

@article{MurArtal2017ORBSLAM2,
  author={Mur-Artal, Ra\'{u}l and Tard\'{o}s, Juan D.},
  journal={IEEE Transactions on Robotics}, 
  title={{ORB-SLAM2: An Open-Source SLAM System for Monocular, Stereo, and RGB-D Cameras}}, 
  year={2017},
  volume={33},
  number={5},
  pages={1255-1262},
  doi={10.1109/TRO.2017.2705103}
}

@book{Lee2000IntroductionToSmoothManifolds,
  author = {Lee, John M.},
  title = {{Introduction to Smooth Manifolds}},
  publisher = {Springer},
  year = {2000}
}

@article{chiu2022SLAMBackends,
      title = {{SLAM Backends with Objects in Motion: A Unifying Framework and Tutorial}}, 
      author={Chiu, Chih-Yuan},
      year={2022},
      eprint={2207.05043},
      booktitle={arXiv},
      primaryClass={cs.RO}
}

\appendix

The ArXiV version of this paper, which contains the appendix, is found here: \url{http://arxiv.org/abs/2207.05043} \cite{chiu2022SLAMBackends}. The author will ensure that the link stays active.

\newpage
~\newpage

The following supplementary material includes the appendix, which contains proofs and figures omitted in the main paper due to space limitations.

\subsection{Proofs of Main Theorems}
\label{subsec: A1, Proofs of Main Theorems}

\begin{theorem} 
\label{Thm: Dynamic EKF, Feature Augmentation}
The feature augmentation step of standard dynamic EKF SLAM (Alg. \ref{Alg: Dynamic EKF, Feature Augmentation, Standard}) is equivalent to applying a Gauss-Newton step to $c_{\dEKF,t,1}: \R^{\dmu + \Nf' \df} \ra \R$, with:
\begin{align*}
    &c_{\dEKF,t,1}(\tilde{x}_t, f_{\nf+\Nf+1}, \cdots, f_{\nf+\Nf+\nf'+\Nf'}) \\
    := \hspace{0.5mm} &\Vert \tilde{x}_t - \mu_t \Vert_{\Sigma_t^{-1}}^2 + \sum_{k=\nf+\Nf+1}^{\Nf + \Nf'} \Vert  z_{t,k} - \hs(\xrt, f_k) \Vert_{\tilde{\Sigma}_v^{-1}}^2.
\end{align*}
\end{theorem}

\begin{proof}
The proof parallels that of \cite{Saxena2022SLAMNonlinearOptimization}, Theorem 5.1, and is reproduced below for completeness.


By assumption, $\frac{\partial h}{\partial f_k}$ is surjective throughout the domain of $h$. Thus, by Theorem \ref{Thm: Implicit Function Theorem, Extension}, given any $x_t \in \R^{\dx}$ and $f_k \in \R^{\df}$, there exists a local \textit{inverse observation map} $\ell: U_x \times U_z \ra U_f$, where $U_x \subset \R^{\dx}$, $U_f \subset \R^{\df}$, and $U_z \subset \R^{\dz}$ are open neighborhoods of $x_t$, $f_k$, and $h(x_t, f_k)$, such that $\hs(\xrt, \ell(\xrt, z_t)) = z_t$ for each $\xrt \in U_x, z_t \in U_z$, and $L_z = \tilde H_{t, f}^\dagger$, where $L_z := \frac{\partial \ell}{\partial z_t} (\xrt, z_t) \in \R^{\df \times \dz}$ and $\tilde H_{t, f} := \frac{\partial h}{\partial f_k}(\xrt, f_k) \in \R^{\dz \times \df}$, and $\dagger$ denotes the Moore-Penrose pseudoinverse. Intuitively, $\ell$ directly generates position estimates of new features from their feature measurements and the current pose, by effectively \say{inverting} the measurement map $h: \R^{\dx} \times \R^{\df} \ra \R^{\dz}$ \cite{Sola2014SLAMWithEKF}. 

First, to simplify notation, define:
\begin{align*}
    z_{t,\text{new}} &= (z_{t,\nf+\Nf+1}, \cdots, z_{t, \text{new}}) \in \R^{(\nf'+\Nf') \dz}, \\
    f_{t,\text{new}} &= (f_{t,\nf+\Nf+1}, \cdots, f_{t \text{new}}) \in \R^{(\nf'+\Nf') \df}, \\
    \tilde{h}(\xrt, f_{t,\text{new}}) &:= \big(\hs(\xrt, f_{t,\nf+\Nf+1}), \cdots, \hs(\xrt, f_{t,\text{new}}) \big) \\
    &\hspace{1cm} \in \R^{(\nf'+\Nf') \dz}, \\
    \tilde{\Sigma}_v &= \text{diag}\{\Sigma_v, \cdots, \Sigma_v \} \in \R^{(\nf'+\Nf')\dz \times (\nf'+\Nf')\dz}.
\end{align*}
We can now rewrite the cost $c_{\dEKF,t,1}$ as:
\begin{align*}
    &c_{\dEKF,t,1}(\tilde{x}_t, f_{t,\text{new}}) \\
    = \hspace{0.5mm} &\Vert \tilde{x}_t - \mu_t \Vert_{\Sigma_t^{-1}}^2 + \Vert z_{t,\text{new}} - \tilde{h}(\xrt, f_{t,\text{new}}) \Vert_{\tilde{\Sigma}_v^{-1}}^2.
\end{align*}

To apply a Gauss-Newton step, we will define $C_1(\tilde{x}_t, f_{t,\text{new}})$ of an appropriate dimension such that $c_{\dEKF,t,1}(\tilde{x}_t, f_{t,\text{new}}) = \Vert C_1(\tilde{x}_t, f_{t,\text{new}}) \Vert_2^2$. A natural choice is furnished by $C_1(\tilde{x}_t, f_{t,\text{new}}) \in \R^{\dx + \Nf \df + \Nf' \dz}$, as defined below:
\begin{align*}
    &C_1(\tilde{x}_t, f_{t,\text{new}}) \\
    := \hspace{0.5mm} &\begin{bmatrix}
        \Sigma_t^{-1/2} (\tilde{x}_t - \mu_t) \\
        \Sigma_v^{-1/2} \big( z_{t,\text{new}} - \tilde{h}(\xrt, f_{t,\text{new}}) \big)
    \end{bmatrix}.
\end{align*}
Thus, our parameters for the Gauss-Newton algorithm submodule are:
\begin{align*}
    \tilde{x}_t^\star &:= ({\xrt}^\star, f_{t,1:\nf+\Nf}^\star, f_{t,\text{new}}^\star) \\
    &= \big( \overline{\mu_t}, \ell({\xrt}^\star, z_{t,\nf+\Nf+1}), \cdots, \ell({\xrt}^\star, z_{t,\text{new}}) \big) \\
    &\hspace{5mm} \in \R^{\dx + (\nf+\Nf+\nf'+\Nf') \df},
\end{align*}
where ${\xrt}^\star \in \R^{\dx}, f_{t,1:\Nf}^\star \in \R^{(\nf+\Nf) \df}, f_{t,\text{new}}^\star \in \R^{(\nf'+\Nf') \df}$, and:
\begin{align*}
    C_1(\tilde{x}_t^\star) &= \begin{bmatrix}
        \Sigma_t^{-1/2} (\tilde{x}_t^\star - \mu_t) \\
        \tilde{\Sigma_v}^{-1/2} \big( z_{t,\text{new}} - \tilde{h}({\xrt}^\star, f_{t,\text{new}}^\star) \big)
    \end{bmatrix} \\
    &= \begin{bmatrix}
    0 \\ 0
    \end{bmatrix} \in \R^{\dmu + (\nf'+\Nf')\dz}, \\
    J &= \begin{bmatrix}
        \Sigma_t^{-1/2} & O \\ - \tilde{\Sigma}_v^{-1/2} \tilde{H}_{t,x} \begin{bmatrix}
        I_{\dx} & O
        \end{bmatrix} & - \tilde{\Sigma}_v^{-1/2} \tilde{H}_{t,f}
    \end{bmatrix} \\
    &\in \R^{(\dmu + (\nf'+\Nf')\dz) \times (\dx + (\nf'+\Nf')\df)},
\end{align*}
where $\tilde{H}_t := \begin{bmatrix}
\tilde{H}_{t,x} & \tilde{H}_{t,f}
\end{bmatrix} \in \R^{(\nf'+\Nf')\dz \times (\dx + (\nf'+\Nf') \df)}$ is defined as the Jacobian of $\tilde{h}: \R^{\dx} \times \R^{(\nf'+\Nf')\df} \ra \R^{(\nf'+\Nf')\dz}$ at $({\xrt}^\star, f_{t,\text{new}}^\star) \in \R^{\dx + (\nf'+\Nf')\df}$, with $\tilde{H}_{t,x} \in \R^{(\nf'+\Nf')\dz \times \dx}$ and $\tilde{H}_{t,f} \in \R^{(\nf'+\Nf')\dz \times (\nf'+\Nf') \df}$. By \cite{Saxena2022SLAMNonlinearOptimization}, Algorithm 4.1, the Gauss-Newton update is thus given by:
{
\normalsize
\begin{align} \nonumber
    &\hspace{5mm} \Sigma_t \\
    &\leftarrow (J^\top J)^{\dagger} \\ \nonumber
    &= \Bigg(\begin{bmatrix}
        \Sigma_t^{-1/2} & - \begin{bmatrix}
        I_{\dx} \\ O
        \end{bmatrix} \tilde{H}_{t,x}^\top \tilde{\Sigma}_v^{-1/2}  \\ O & - \tilde{\Sigma}_v^{-1/2} \tilde{H}_{t,f}
    \end{bmatrix} \\
    &\hspace{5mm} \cdot
    \begin{bmatrix}
        \Sigma_t^{-1/2} & O \\ - \tilde{\Sigma}_v^{-1/2} \tilde{H}_{t,x} \begin{bmatrix}
        I_{\dx} & O
        \end{bmatrix} & - \tilde{\Sigma}_v^{-1/2} \tilde{H}_{t,f}
    \end{bmatrix} \Bigg)^{\dagger} \\ 
    \label{Eqn: EKF SLAM Proof, Feature Aug, J T J inverse}
    &= \begin{bmatrix}
        \Omega_{t,xx} + \tilde{H}_{t,x}^\top \tilde{\Sigma}_v^{-1} \tilde{H}_{t,x} & \Omega_{t,xe} & \tilde{H}_{t,x}^\top \tilde{\Sigma}_v^{-1} \tilde{H}_{t,f} \\
        \Omega_{t,ex} & \Omega_{t,ee} & O \\
        \tilde{H}_{t,f}^\top \tilde{\Sigma}_v^{-1} \tilde{H}_{t,x} & O & \tilde{H}_{t,f}^\top \tilde{\Sigma}_v^{-1} \tilde{H}_{t,f}
    \end{bmatrix}^{\dagger}, \\ \nonumber
    \overline{\mu_t} &\gets \tilde{x}_t^\star - (J^\top J)^{\dagger} J^\top C_1(\tilde{x}_t^\star) \\ \nonumber
    &= \big( \overline{\mu_t}, \ell({\xrt}^\star, z_{t,\nf+\Nf+1}), \cdots, \ell({\xrt}^\star, z_{t,\text{new}}) \big),
\end{align}}
where $\dagger$ denotes the Moore-Penrose pseudoinverse.

Here, we have defined $\Omega_{t,xx} \in \R^{\dx \times \dx}, \Omega_{t,xe} = \Omega_{t,ex}^\top$ and $\Omega_{t,ee}$ by:
\begin{align} 
\label{Eqn: EKF SLAM proof, Omega def}
     &\begin{bmatrix}
         \Omega_{t,xx}  & \Omega_{t,xe} \\
        \Omega_{t,ex} & \Omega_{t,ee}
     \end{bmatrix} := \begin{bmatrix}
         \Sigma_{t,xx}  & \Sigma_{t,xe} \\
        \Sigma_{t,ex} & \Sigma_{t,ee}
     \end{bmatrix}^{-1} \\ \nonumber
     &\hspace{5mm} \in \R^{(\dmu + (\nf'+\Nf') \df) \times (\dmu + (\nf'+\Nf') \df)}
\end{align}

To conclude the proof, we must show that \eqref{Eqn: EKF SLAM Proof, Feature Aug, J T J inverse} is identical to the update equations for covariance matrix in the standard formulation of the Extended Kalman Filter algorithm, i.e., we must show that:
\begin{align*}
    &\begin{bmatrix}
    \Sigma_{t,xx} & \Sigma_{t,xe} & \Sigma_{t,xx} L_x^\top \\
    \Sigma_{t,ex} & \Sigma_{t,ee} & \Sigma_{t,ex} L_x^\top \\
    L_x \Sigma_{t,xx} & L_x \Sigma_{t,xe} & L_x \Sigma_{t,xx} L_x^\top + L_z \Sigma_v L_z^\top
    \end{bmatrix} \\
    &\hspace{1cm} = \begin{bmatrix}
        \Omega_{t,xx} + \tilde{H}_{t,x}^\top \tilde{\Sigma}_v^{-1} \tilde{H}_{t,x} & \Omega_{t,xe} & \tilde{H}_{t,x}^\top \tilde{\Sigma}_v^{-1} \tilde{H}_{t,f} \\
        \Omega_{t,ex} & \Omega_{t,ee} & O \\
        \tilde{H}_{t,f}^\top \tilde{\Sigma}_v^{-1} \tilde{H}_{t,x} & O & \tilde{H}_{t,f}^\top \tilde{\Sigma}_v^{-1} \tilde{H}_{t,f}
    \end{bmatrix}^{\dagger}
\end{align*}
This follows by applying \eqref{Eqn: EKF SLAM proof, Omega def}, the fact that $L_z = \tilde H_{t, f}^\dagger$, as well as the matrix equalities resulting from taking the derivative of the equation $z_t := h \big(\xrt, \ell(\xrt, z_t) \big)$ with respect to $\xrt \in \R^{\dx}$ and $z_t \in \R^{\dz}$, respectively:
\begin{align*}
    I_{\dz} &= \tilde{H}_{t,f} L_z, \\
    O &= \tilde{H}_{t,x} + H_{t,f} L_x.
\end{align*}
\end{proof}

Below, we present the proof of Theorem \ref{Thm: Dynamic EKF, Dynamic Object Pose Augmentation}, restated below.

\begin{theorem} 
The dynamic object pose augmentation step of standard dynamic EKF SLAM (Alg. \ref{Alg: Dynamic EKF, Moving Object Pose Augmentation, Standard}) is equivalent to applying a Gauss-Newton step to $c_{\dEKF,t,3}: \R^{\dmu + \no \dx} \ra \R$, with:
\begin{align*}
    &c_{\dEKF,t,3}(\tilde{x}_t, \fpose_{t, 1}, \cdots, \fpose_{t, \no}) \\
    := \hspace{0.5mm} &\Vert \tilde{x}_t - \mu_t \Vert_{\Sigma_t^{-1}}^2 + \sum_{\alpha=1}^{\no} \sum_{k=1}^{\nof(\alpha)} \Vert \fd_{t, \alpha, k} - \go (\fpose_{t, \alpha}, \fd_{0, \alpha, k}) \Vert_{\Sigma_\fpose^{-1}}^2.
\end{align*}
when $\Sigma_{\fpose}$ is a diagonal matrix.
\end{theorem}

\begin{proof}

Define, for convenience:
\begin{align*}
    \fd_{\tau, \alpha} &:= (\fd_{\tau, \alpha, 1}, \cdots, \fd_{\tau, \alpha, \nof(\alpha)}) \in \R^{\nof(\alpha) \df}, \\
    &\hspace{1cm} \tau = 0, t, \\
    \Nf &:= \sum_{\alpha=1}^{\no} \nof(\alpha) \in \N, \\
    \fd_t &:= (\fd_{t, 1}, \cdots, \fd_{t, \no}) \in \R^{\Nf \df},  \hspace{5mm} \tau = 0, t, \\
    \tilde g^o(\fpose_{t}, \fd_{0}) &:= \big( g^o(\fpose_{t}, \fd_{0, 1, 1}), \cdots, g^o(\fpose_{t}, \fd_{0, 1, \nof(1)}), \cdots, \\
    &\hspace{1cm} g^o(\fpose_{t}, \fd_{0, \no, 1}), \cdots, g^o(\fpose_{t}, \fd_{0, \no, \nof(\no)}) \big) \\
    &\hspace{1.5cm} \in \R^{\Nf \df}, \\
    \tilde \Sigma_{\fpose} &:= diag\{\Sigma_{\fpose}, \cdots, \Sigma_{\fpose} \} \in \R^{\no \dx \times \no \dx}, \\
    \tilde x_t &:= (\tilde \xr_{t, e}, \fd_{0}, \fd_t) \in \R^{\dmu},
\end{align*}
where the final definition follows from the variable reordering process described in Alg. \ref{Alg: Dynamic EKF, Moving Object Pose Augmentation, Standard}, Line \ref{Eqn: Dynamic EKF, Moving Object Pose Augmentation, Variable Reordering, 1}.

Let $\tilde \invg: \R^{\Nf \df} \times \R^{\Nf \df} \ra \R^{\no \dx}$ 
be given such that $\tilde \invg (\fd_0, \go(\fpose_t, \fd_0)) = \fpose_t$,
for any $\fpose_t \in \R^{\dx}, \fd_0 \in \R^{\df}$, and such that $\tilde \Invg_t = \tilde {\Gxio}^\dagger$. The cost $c_{\dEKF, t, 3}: \R^{\dmu} \times \R^{\no \dx} \ra \R$ can now be written as:
\begin{align*}
    &c_{\dEKF, t, 3}(\tilde x_t, \fpose_{t}) \\
    = \hspace{0.5mm} &\Vert \tilde x_t - \mu_t \Vert_{\Sigma_t^{-1}}^2 + \Vert \fd_t - \tilde \go(\fpose_{t}, \fd_{0}) \Vert_{\Sigma_\fpose^{-1}}^2
\end{align*}
To apply a Gauss-Newton step, define:
\begin{align*}  
    &C_2(\tilde x_t, \fpose_{t}) := \begin{bmatrix}
        \Sigma_t^{-1/2} (\tilde x_t - \mu_t) \\
        \tilde \Sigma_\fpose^{-1/2} (\fd_t - \tilde \go(\fpose_{t}, \fd_{0}))
    \end{bmatrix} \\
    &\hspace{5mm} \in \R^{(\dmu + \Nf \df) \times \dmu}
\end{align*}
such that $c_{\dEKF, t, 2}(\tilde x_t, \fpose_{t}) = \Vert C_2(\tilde x_t, \fpose_{t}) \Vert_{\Sigma_\fpose^{-1}}^2$. The Gauss-Newton algorithm submodule thus has parameters:
\begin{align*}
    ({\xrt}^\star, \fpose_{t}^\star)  &:= \big( \mu_t, \tilde \invg(\fd_{0}, \fd_t) \big) \in \R^{\dmu + \no \dx}, \\
    C_2(\tilde {\xrt}^\star, \fpose_{t}^\star) &= \begin{bmatrix}
        \Sigma_t^{-1/2} (\tilde {\xrt}^\star - \mu_t) \\
        \tilde \Sigma_\fpose^{-1/2} (\fd_t - \tilde \go(\fpose_{t}^\star, \fd_{0}))
    \end{bmatrix} 
    = \begin{bmatrix}
        0 \\ 0
    \end{bmatrix} \\
    &\hspace{1cm} \in \R^{\dmu + \Nf \df}, \\
    J &= \begin{bmatrix}
        \Sigma_t^{-1/2} & O \\
         - \tilde \Sigma_{\fpose}^{-1/2} \begin{bmatrix}
            O & - \tilde \Gfo & I_{\df}
        \end{bmatrix} & - \tilde \Sigma_{\fpose}^{-1/2} \tilde \Gfo
    \end{bmatrix} \\
    &\hspace{1cm} \in \R^{(\dmu + \Nf \df) \times (\dmu + \no \dx)}.
\end{align*}
We thus have:
{\footnotesize
\begin{align*} 
    &\hspace{5mm} J^\top J \\
    &= 
    \begin{bmatrix}
        \Sigma_t^{-1/2} &
         - \begin{bmatrix}
            O \\ - \tilde \Gfo^\top \\ I_{\df}
        \end{bmatrix} \tilde \Sigma_{\fpose}^{-1/2} \\
        O & - \tilde \Gfo^\top \tilde \Sigma_{\fpose}^{-1/2}
    \end{bmatrix} \\
    &\hspace{5mm} \cdot
    \begin{bmatrix}
        \Sigma_t^{-1/2} & O \\
         - \tilde \Sigma_{\fpose}^{-1/2}
         \begin{bmatrix}
            O & - \tilde \Gfo & I_{\df}
        \end{bmatrix} & - \tilde \Sigma_{\fpose}^{-1/2} \tilde \Gfo
    \end{bmatrix} \\
    &= \begin{bmatrix}
    \Sigma_t^{-1} + \begin{bmatrix}
            O \\ - \tilde \Gfo^\top \\ I_{\df}
        \end{bmatrix} \tilde \Sigma_{\fpose}^{-1/2} 
        \begin{bmatrix}
            O & - \tilde \Gfo^\top & I_{\df}
        \end{bmatrix} & \begin{bmatrix}
            O \\ - \tilde \Gfo^\top \\ I_{\df}
        \end{bmatrix} \tilde \Sigma_{\fpose}^{-1} \tilde \Gxio \\
        {\Gxio}^\top \tilde \Sigma_{\fpose}^{-1}  \begin{bmatrix}
            O & - \tilde \Gfo & I_{\df}
        \end{bmatrix} & \tilde {\Gxio}^\top \tilde \Sigma_{\fpose}^{-1} \tilde \Gxio
    \end{bmatrix} \\
    &= \begin{bmatrix}
    \Omega_{t, ee} & \Omega_{t, ef} & \Omega_{t, ef'} & O \\
    \Omega_{t, fe} & \Omega_{t, ff} + \tilde {G_f^o}^\top \tilde \Sigma_{\fpose}^{-1} \tilde G_f^o & \Omega_{t, ff'} - \tilde {G_f^o}^\top \tilde \Sigma_{\fpose}^{-1} & \tilde \Gfo^\top \tilde \Sigma_{\fpose}^{-1} \tilde \Gxio \\
    \Omega_{t, f'e} & \Omega_{t, f'f} - \tilde \Sigma_{\fpose}^{-1} \tilde G_f^o & \Omega_{t, f'f'} + \tilde \Sigma_{\fpose}^{-1`} & - \tilde \Sigma_{\fpose}^{-1} \tilde \Gxio \\
    O & \tilde {\Gxio}^\top \tilde \Sigma_{\fpose}^{-1} \tilde \Gfo & - \tilde {\Gxio}^\top \tilde \Sigma_{\fpose}^{-1} & \tilde {\Gxio}^\top \tilde \Sigma_{\fpose}^{-1} \Gxio
    \end{bmatrix},
\end{align*}
}
where we have defined:
\begin{align*}
    \begin{bmatrix}
    \Omega_{t, ee} & \Omega_{t, ef} & \Omega_{t, ef'} \\
    \Omega_{t, fe} & \Omega_{ff} & \Omega_{ff'} \\
    \Omega_{t, f'e} & \Omega_{f'f} & \Omega_{f'f'}
    \end{bmatrix} 
    := \Sigma_t^{-1} \in \R^{\dmu \times \dmu},
\end{align*}
with $\Omega_{t, ff}, \Omega_{t, f'f'} \in \R^{\Nf \df \times \Nf \df}$. The Gauss-Newton update (\cite{Saxena2022SLAMNonlinearOptimization}, Algorithm 3) now gives:
\begin{align*}
    \Sigma_t &\gets (J^\top J)^\dagger \\
    &= \begin{bmatrix}
        \Sigma_{t, ee} & \Sigma_{t, ef} & \Sigma_{t, ef'} & \Sigma_{t, \fpose e}^\top \\
        \Sigma_{t, fe} & \Sigma_{t, ff} & \Sigma_{t, ff'} & \Sigma_{t, \fpose f}^\top \\
        \Sigma_{t, f'e} & \Sigma_{t, f'f} & \Sigma_{t, f'f'} & \Sigma_{t, \fpose f'}^\top \\
        \Sigma_{t, \fpose e} & \Sigma_{t, \fpose f} & \Sigma_{t, \fpose f'} & \Sigma_{t, \fpose \fpose}
    \end{bmatrix} \in \R^{\dmu \times \dmu}
\end{align*}
where:
\begin{align*}
    \begin{bmatrix}
        \Sigma_{t, ee} & \Sigma_{t, ef} & \Sigma_{t, ef'} \\
        \Sigma_{t, fe} & \Sigma_{t, ff} & \Sigma_{t, ff'} \\
        \Sigma_{t, f'e} & \Sigma_{t, f'f} & \Sigma_{t, f'f'} 
    \end{bmatrix} = \Sigma_t,
\end{align*}
with $\Sigma_{t, ff}, \Sigma_{t, f'f'} \in \R^{\Nf \df \times \Nf \df}$, and:
\begin{align*}
    \Sigma_{t, \fpose e} &:= \tilde \Invg_1 \Sigma_{t, fe} + \tilde \Invg_2 \Sigma_{t, f' e} \in \R^{\no \dx \times (\dmu - 2 \Nf \df)}, \\
    \Sigma_{t, \fpose f} &:= \tilde \Invg_1 \Sigma_{t, ff} + \tilde \Invg_2 \Sigma_{t, f' f} \in \R^{\no \dx \times \Nf \df}, \\
    \Sigma_{t, \fpose f'} &:= \tilde \Invg_1 \Sigma_{t, ff'} + \tilde \Invg_2 \Sigma_{t, f' f'} \in \R^{\no \dx \times \Nf \df}, \\
    \Sigma_{t, \fpose \fpose} &:= \tilde \Invg_1 \Sigma_{t, ff} \tilde \Invg_1^\top + \tilde \Invg_2 \Sigma_{t, f' f} \tilde \Invg_1^\top + \tilde \Invg_1 \Sigma_{t, f f'} \tilde \Invg_2^\top
    \\
    &\hspace{1cm} + \tilde \Invg_2 (\Sigma_{t, f' f'} + \tilde \Sigma_\fpose) \tilde \Invg_2^\top \in \R^{\no \dx \times \no \dx}
\end{align*}

\end{proof}

\begin{theorem} 
\label{Thm: Dynamic EKF, Static Feature Update}
The static feature update step of standard dynamic EKF SLAM (Alg. \ref{Alg: Dynamic EKF, Static Feature Update, Standard}) is equivalent to applying a Gauss-Newton step to $c_{\dEKF,t,5}: \R^{\dmu + \Nf' \df} \ra \R$, with:
\begin{align*}
    &c_{\dEKF,t,5}(\tilde{x}_t, f_{\nf+1}, \cdots, f_{\nf+\nf'}) \\
    := \hspace{0.5mm} &\Vert \tilde{x}_t - \mu_t \Vert_{\Sigma_t^{-1}}^2 + \sum_{k=1}^{\nf} \|\zs_{t,k} - \hs(\xrt, \fs_k)\|_{\Sigma_{v}^{-1}}^2.
\end{align*}
\end{theorem}

\begin{proof}
The proof parallels that of \cite{Saxena2022SLAMNonlinearOptimization}, Theorem 5.2, and is reproduced below for completeness.

First, to simplify notation, define:
\begin{align*}
    z_t &:= (z_{t,1}, \cdots, z_{t,\nf}) \in \R^{\nf \dz}, \\
    \fs &:= (\fs_{1}, \cdots, \fs_{\nf}) \in \R^{\nf \df}, \\
    \Tilde{h}(\xrt, \fs) &:= \big( \hs(\xrt, \fs_{1}), \cdots, \hs(\xrt, \fs_{\nf}) \big) \in \R^{\nf \dz}, \\
    \Tilde{\Sigma}_v &:= \text{diag}\{\Sigma_v, \cdots, \Sigma_v \} \in \R^{\nf \dz \times \nf \dz}.
\end{align*}
We can then rewrite the cost as:
\begin{align*}
    c_{\dEKF,t,5}(\tilde{x}_t) = 
    \Vert \tilde{x}_t^\star - \mu_t \Vert_{\Sigma_t^{-1}}^2 + \Vert z_t - \tilde{h}(\tilde{x}_t^\star)  \Vert_{\tilde{\Sigma}_v^{-1}}^2.
\end{align*}

To apply a Gauss-Newton step, we seek a vector $C_3(\tilde{x}_t)$ of an appropriate dimension such that $c_{\dEKF,t,2}(\tilde{x}_t) = C_3(\tilde{x}_t)^\top C_3(\tilde{x}_t)$. A natural choice is furnished by $C_3(\tilde{x}_t) \in \R^{\dmu + \nf \dz}$, as defined below:
\begin{align*}
    C_3(\tilde{x}_t) := \begin{bmatrix} 
    \Sigma_t^{-1/2}(\tilde{x}_t - \mu_t) \\ \tilde{\Sigma}_v^{-1/2}(z_t - \Tilde{h}(\tilde{x}_t)) \end{bmatrix}.
\end{align*}
Thus, our parameters for the Gauss-Newton algorithm submodule are:
\begin{align*}
    \tilde{x}_t^\star &= \mu_t \in \R^{\dmu}, \\
    C_3(\tilde{x}_t^\star) &= 
    \begin{bmatrix} 
    \Sigma_
    t^{-1/2}(\tilde{x}_t^\star - \mu_t) \\ \tilde{\Sigma}_v^{-1/2}(z_t - \Tilde{h}(\tilde{x}_t^\star)) \end{bmatrix} = \begin{bmatrix} 
    0 \\ \tilde{\Sigma}_v^{-1/2} \big(z_t - \tilde{h}(\mu_t) \big) \end{bmatrix} \\
    &\in \R^{\dmu + \nf \dz}, \\
    J &= \begin{bmatrix}
        \Sigma_t^{-1/2} \\
        -\tilde{\Sigma}_v^{-1/2} \begin{bmatrix}
            H_t & O
        \end{bmatrix} 
    \end{bmatrix} \in \R^{(\dmu + \nf \dz) \times \dmu},
\end{align*}
where $\tilde{H}_t \in \R^{\nf \dz} \times \R^{\dx + \nf \df}$ is defined as the Jacobian of $\Tilde{h}: \R^{\dx} \times \R^{\nf \df} \ra \R^{\nf \dz}$ at $\tilde{x}_t^\star \in \R^{\dmu}$. By \cite{Saxena2022SLAMNonlinearOptimization}, Algorithm 4.1, the Gauss-Newton update is thus given by:
    \begin{align*}
        \overline{\Sigma}_t &\leftarrow (J^\top J)^{-1} \\
        &= \left(\Sigma_t^{-1} + \begin{bmatrix}
            H_t^\top \\ O
        \end{bmatrix} \tilde \Sigma_v \begin{bmatrix}
            H_t & O
        \end{bmatrix} \right)^{-1} \\
        &= \Sigma_t - \Sigma_t \begin{bmatrix}
            H_t^\top \\ O
        \end{bmatrix} \left(\tilde \Sigma_v^{-1} + \begin{bmatrix}
            H_t & O
        \end{bmatrix} \Sigma_t \begin{bmatrix}
            H_t^\top \\ O
        \end{bmatrix} \right)^{-1} \\
        &\hspace{1cm} \cdot
        \begin{bmatrix}
            H_t & O
        \end{bmatrix}
        \Sigma_t, \\
        \overline{\mu_t} &\gets \mu_t - (J^\top J)^{-1} J^\top C_3(\tilde{x}_t^\star) \\
        &f= \mu_t - (\Sigma^{-1}_t + H_t^\top \tilde \Sigma_v^{-1} H_t)^{-1} \begin{bmatrix}
        \Sigma_t^{-1/2} & - H_t^\top \tilde \Sigma_v^{-1/2}
        \end{bmatrix} \\
        &\hspace{2cm} \cdot
        \begin{bmatrix}
        0 \\ \tilde \Sigma_v^{-1/2}(z_t - \tilde{h}(\mu_t))
        \end{bmatrix} \\
        &= \mu_t + (\Sigma^{-1}_t + H_t^\top \tilde \Sigma_v^{-1} H_t)^{-1} H_t^\top \tilde \Sigma_v^{-1} \big( z_t - \tilde{h}(\mu_t) \big),\\
        &= \mu_t + \tilde \Sigma_v^{-1} \begin{bmatrix}
            H_t^\top \\ O
        \end{bmatrix} \left( \begin{bmatrix}
            H_t & O
        \end{bmatrix} \Sigma_t
        \begin{bmatrix}
            H_t^\top \\ O
        \end{bmatrix} + \tilde \Sigma_v \right)^{-1} \\
        &\hspace{1cm} \cdot \big( z_t - \tilde{h}(\mu_t) \big),
    \end{align*}
    which are identical to the feature update equations for the mean and covariance matrix in the Extended Kalman Filter algorithm, i.e. \eqref{Eqn: Dynamic EKF, Mean Update} and \eqref{Eqn: Dynamic EKF, Cov Update} respectively. Note that, in the final step, we have used a variant of the Woodbury Matrix Identity.
\end{proof}

Below, we present the proof of Theorem \ref{Thm: Dynamic EKF, Smoothing Update}, restated below.

\begin{theorem} 
The smoothing update step of standard dynamic EKF SLAM (Alg. \ref{Alg: Dynamic EKF, Smoothing Update, Standard}) is equivalent to applying a Gauss-Newton step to $c_{\dEKF,t,7}: \R^{\dmu} \ra \R$, with:
\begin{align*}
    &c_{\dEKF,t,7}(\tilde{x}_t) \\
    := \hspace{0.5mm} &\Vert \tilde{x}_t - \mu_t \Vert_{\Sigma_t^{-1}}^2 + \sum_{\alpha=1}^{\no} \Vert s(\fpose_{t-2, \alpha}, \fpose_{t-1, \alpha}, \fpose_{t, \alpha}) \Vert_{\Sigma_s^{-1}}^2.
\end{align*}
\end{theorem}

\begin{proof}

To simplify notation, define:
\begin{align*}
    \fpose_{\tau} &:= (\fpose_{\tau, 1}, \cdots, \fpose_{\tau, \no}) \in \R^{\no \dx}, \hspace{1cm} \forall \hspace{0.5mm} \tau \in \{t-2, t-1, t\}, \\
    \tilde \Sigma_s &:= diag\{\Sigma_s, \cdots, \Sigma_s \} \in \R^{\no \dx \times \no \dx}.
\end{align*}
We can then rewrite the cost as:
\begin{align*}
    &c_{\dEKF, t, 7}(\tilde x_t) \\
    := \hspace{0.5mm} &\Vert \tilde{x}_t - \mu_t \Vert_{\Sigma_t^{-1}}^2 + \Vert s(\fpose_{t-2}, \fpose_{t-1}, \fpose_{t}) \Vert_{\tilde \Sigma_s^{-1}}^2.
\end{align*}

To apply a Gauss-Newton step, we must find a vector $C_4(\tilde x_t)$, of appropriate dimension, such that $c_{\dEKF, t, 7} = \Vert C_4(\tilde x_t) \Vert_2^2$. To this end, define:
\begin{align*}
    C_4(\tilde x_t) &:= \begin{bmatrix}
        \Sigma_t^{-1/2} (\tilde x_t - \overline{\mu_t}) \\
        \tilde \Sigma_s^{-1/2} s(\fpose_{t-2}, \fpose_{t-1}, \fpose_{t})
    \end{bmatrix} \in \R^{\dmu + \dx}.
\end{align*}
The Gauss-Newton submodule parameters are therefore:
\begin{align*}
    \tilde{x}_t^\star &= \mu_t \in \R^{\dx + \nf \df}, \\
    C_4(\tilde{x}_t^\star) &= 
     \begin{bmatrix} 
        \Sigma_t^{-1/2} (\tilde {\xrt}^\star - \overline{\mu_t}) \\
        s(\fpose_{t-2}, \fpose_{t-1}, \fpose_{t})
    \end{bmatrix} \\
    &\in \R^{\dx + \nf \df + \nf \dz}, \\
    J &= \begin{bmatrix}
        \Sigma_t^{-1/2} \\
        \begin{matrix}
            O & \tilde \Sigma_s^{-1/2} S_{t-2} & \tilde \Sigma_s^{-1/2} S_{t-1} & \tilde \Sigma_s^{-1/2} S_{t}
        \end{matrix}
    \end{bmatrix} \\
    &\hspace{1cm} \cdot \in \R^{(\dmu + \dx) \times \dmu},
\end{align*}
and thus:
\begin{align*}
    J^\top J &= \begin{bmatrix}
        \Sigma_t^{-1/2} & \begin{matrix}
        O \\ S_{t-2}^\top \tilde \Sigma_s^{-1/2} \\ S_{t-1}^\top \tilde \Sigma_s^{-1/2} \\ S_{t}^\top \tilde \Sigma_s^{-1/2}
        \end{matrix}
    \end{bmatrix} \\
    &\hspace{1cm} \cdot
    \begin{bmatrix}
        \Sigma_t^{-1/2} \\
        \begin{matrix}
            O & \tilde \Sigma_s^{-1/2} S_{t-2} & \tilde \Sigma_s^{-1/2} S_{t-1} & \tilde \Sigma_s^{-1/2} S_{t}
        \end{matrix}
    \end{bmatrix} \\
    &= \Sigma_t^{-1} + 
    \begin{bmatrix}
         O \\ S_{t-2}^\top \\ S_{t-1}^\top \\ S_{t}^\top
    \end{bmatrix} \tilde \Sigma_s^{-1}
    \begin{bmatrix}
        O & S_{t-2} & S_{t-1} & S_{t}
    \end{bmatrix}
\end{align*}
By \cite{Saxena2022SLAMNonlinearOptimization}, Algorithm 4.1, the Gauss-Newton update is given by:
{\small
\begin{align*}
    \Sigma_t &\gets (J^\top J)^{-1} \\
    &= \left( \Sigma_t^{-1} + 
    \begin{bmatrix}
         O \\ S_{t-2}^\top \\ S_{t-1}^\top \\ S_{t}^\top
    \end{bmatrix} \tilde \Sigma_s^{-1}
    \begin{bmatrix}
        O & S_{t-2} & S_{t-1} & S_{t}
    \end{bmatrix} \right)^{-1} \\
    &= \Sigma_t - \Sigma_t \begin{bmatrix}
         O \\ S_{t-2}^\top \\ S_{t-1}^\top \\ S_{t}^\top
    \end{bmatrix} \\
    &\hspace{1cm} \cdot
    \left( \tilde \Sigma_s + 
    \begin{bmatrix}
        O & S_{t-2} & S_{t-1} & S_{t}
    \end{bmatrix} \Sigma_t \begin{bmatrix}
         O \\ S_{t-2}^\top \\ S_{t-1}^\top \\ S_{t}^\top
    \end{bmatrix}
    \right)^{-1} \\
    &\hspace{1cm} \cdot \begin{bmatrix}
        O & S_{t-2} & S_{t-1} & S_{t}
    \end{bmatrix} \Sigma_t \\
    \mu_t &\gets \mu_t - (J^\top J)^{-1} J^\top C_4(\mu_t) \\
    &= \mu_t - \left( \Sigma_t^{-1} + 
    \begin{bmatrix}
        O \\ S_{t-2}^\top \\ S_{t-1}^\top \\ S_{t}^\top
    \end{bmatrix}
    \tilde \Sigma_s^{-1}
    \begin{bmatrix}
        O & S_{t-2} & S_{t-1} & S_{t}
    \end{bmatrix} \right)^{-1} \\
    &\hspace{1cm}
    \begin{bmatrix}
        \Sigma_t^{-1/2} & 
        \begin{matrix}
            O \\ S_{t-2}^\top \tilde \Sigma_s^{-1/2} \\ S_{t-1}^\top \tilde \Sigma_s^{-1/2} \\ S_{t}^\top \tilde \Sigma_s^{-1/2}
        \end{matrix}
    \end{bmatrix} \cdot 
    \begin{bmatrix}
        O \\ \tilde \Sigma_s^{-1/2} s(\fpose_{t-2}, \fpose_{t-1}, \fpose_{t}) 
    \end{bmatrix} \\
   &= \mu_t - \left( \Sigma_t^{-1} + 
    \begin{bmatrix}
        O \\ S_{t-2}^\top \\ S_{t-1}^\top \\ S_{t}^\top
    \end{bmatrix}
    \tilde \Sigma_s^{-1}
    \begin{bmatrix}
        O & S_{t-2} & S_{t-1} & S_{t}
    \end{bmatrix} \right)^{-1} \\
    &\hspace{1cm} \cdot 
    \begin{bmatrix}
        O \\ S_{t-2}^\top \\ S_{t-1}^\top \\ S_{t}^\top
    \end{bmatrix} \tilde \Sigma_s^{-1} s(\fpose_{t-2}, \fpose_{t-1}, \fpose_{t}) \\
    &= \mu_t - \Sigma_t \begin{bmatrix}
        O \\ S_{t-2}^\top \\ S_{t-1}^\top \\ S_{t}^\top
    \end{bmatrix} \\
    &\hspace{1cm} \cdot \left( \tilde \Sigma_s + 
    \begin{bmatrix}
        O & S_{t-2} & S_{t-1} & S_{t}
    \end{bmatrix} \Sigma_t 
    \begin{bmatrix}
        O \\ S_{t-2}^\top \\ S_{t-1}^\top \\ S_{t}^\top
    \end{bmatrix} \right)^{-1} \\
    &\hspace{1cm} \cdot s(\fpose_{t-2}, \fpose_{t-1}, \fpose_{t})
\end{align*}
}
via the Woodbury Matrix Identity. This establishes the equivalence between the smoothing update equations for the dynamic EKF algorithm as written in our optimization-based algorithm, and as written in the mold of the standard EKF SLAM algorithm \cite{Thrun2005ProbabilisticRobotics}.

\end{proof}

\begin{theorem} 
\label{Thm: Dynamic EKF, State Propagation}
The state propagation step of standard dynamic EKF SLAM (Alg. \ref{Alg: Dynamic EKF, State Propagation, Standard}) is equivalent to applying a Gauss-Newton step to $c_{\dEKF,t,9}: \R^{\dmu} \ra \R$, with:
\begin{align*}
    &c_{\dEKF,t,9}(\tilde{x}_t, x_{t+1}) \\
    := \hspace{0.5mm} &\Vert \tilde{x}_t - \overline \mu_t \Vert_{\Sigma_t^{-1}}^2 + \Vert \xr_{t+1} - g(\xrt) \Vert_{\Sigma_w^{-1}}^2.
\end{align*}
\end{theorem}

\begin{proof}   
The proof parallels that of \cite{Saxena2022SLAMNonlinearOptimization}, Theorem 5.3, and is reproduced below for completeness.

Intuitively, the state propagation step marginalizes out $x_t \in \R^{\dx}$ in the full state vector $\tilde x_t \in \R^{\dx}$, and retains $\xr_{t+1} \in \R^{\dx}$. In the notation of our Marginalization algorithm submodule:
\begin{align*}
    \tilde{x}_{t,K} &= (x_{t+1}, \fs, \fd, \fpose) \in \R^{\dmu}, \\
    \tilde{x}_{t,M} &= \xrt \in \R^{\dx},
\end{align*}
with the costs:
\begin{align*}
    c_{K}(\xr_{t+1}) &= 0 \\
    c_{M}(\tilde{x}_t, \xr_{t+1}) &= \Vert \tilde{x}_t - \overline{\mu_t} \Vert_{\overline{\Sigma}_t^{-1}}^2 + \Vert \xr_{t+1}-g(\xrt) \Vert_{\Sigma_w^{-1}}^2.
\end{align*}
To apply a marginalization step, we must first define vectors $C_{K}(\xr_K) = C_{K}(\tilde{x}_t)$ and $C_{M}(\xr_K, \xr_M) = C_{M}(\tilde{x}_t, \xr_{t+1})$ of appropriate dimensions such that $c_{\dEKF,t,9}(\tilde{x}_t, \xr_{t+1}) = C_{K}(\xr_{t+1})^\top C_{K}(\xr_{t+1}) + C_{M}(\tilde{x}_t, \xr_{t+1})^\top C_{M}(\tilde{x}_t, \xr_{t+1})$. To this end, we identify the following parameters, in the language of a Marginalization step (\cite{Saxena2022SLAMNonlinearOptimization}, Section 2):
\begin{align*}
    C_{K}(\tilde{x}_{t,K}) &:= 0 \in \R \\
    C_{M}(\tilde{x}_{t,K},\tilde{x}_{t,M}) &:= \begin{bmatrix} \bar\Sigma_t^{-1/2}(\tilde{x}_t - \overline{\mu_t}) \\ \Sigma_w^{-1/2} \big(\xr_{t+1} - g(\xrt) \big) \end{bmatrix} \in \R^{\dmu + \dx}.
\end{align*}
For convenience, we will define the pose and feature track components of the mean $\mu_t \in \R^{\dmu}$ by $\mu_t := (\mu_{t,x}, \mu_{t,e}) \in \R^{\dmu}$, with $\mu_{t,x} \in \R^{\dx}$. 
In addition, we will define the components of $\bar\Sigma_t^{-1/2} \in \R^{\dmu \times \dmu}$ and $\bar\Sigma_t^{-1} \in \R^{\dmu \times \dmu}$ by:
\begin{align*}
   \begin{bmatrix}
        \Omega_{t,xx} & \Omega_{t,xe} \\
        \Omega_{t,ex} & \Omega_{t,ee}
    \end{bmatrix} &:=  \bar\Sigma_t^{-1} \in \R^{\dmu \times \dmu}, \\
    \begin{bmatrix}
        \Invg_{t,xx} & \Invg_{t,xe} \\
        \Invg_{t,ex} & \Invg_{t,ee}
    \end{bmatrix} &:= \bar\Sigma_t^{-1/2} \in \R^{\dmu \times \dmu},
\end{align*}
where $\Sigma_{t,xx}, \Invg_{t,xx} \in \R^{\dx \times \dx}$. Using the above definitions, we can reorder the residuals in $C_K \in \R$ and $C_M \in \R^{\dmu + \dx}$, and thus redefine them by:
\begin{align*}
    &C_{K}(\tilde{x}_{t,K}) =0 \in \R \\
    &C_{M}(\tilde{x}_{t,K},\tilde{x}_{t,M}) \\
    &= \begin{bmatrix} 
        \Invg_{t,xx}(\xrt - \mu_{t,x}) + \Invg_{t,xe}(\fs - \mu_{t,f}) \\
        \Sigma_w^{-1/2}(\xr_{t+1} - g(\xrt)) \\
        \Invg_{t,ex}(\xrt - \mu_{t,x}) + \Invg_{t,ee}(\fs - \mu_{t,f})
    \end{bmatrix} \in \R^{\dmu + \dx}.
\end{align*}
Our state variables and cost functions for the Gauss-Newton algorithm submodule are:
\begin{align*}
    \overline{{\xr_M}^\star} &= \tilde{x}_t^\star = \overline{\mu_t} \in \R^{\dmu}, \\
    \overline{{\xr_K}^\star} &= g(\tilde{x}_t^\star) = g(\overline{\mu_t}) \in \R^{\dmu}, \\
    C_{K}(\tilde{x}_{t,K}^\star) &= 0 \in \R, \\
    C_{M}(\tilde{x}_{t,K}^\star,\tilde{x}_{t,M}^\star) &= 
    \begin{bmatrix}
    0 \\ 0
    \end{bmatrix} \in \R^{\dmu + \dx}, \\
    J_K &= \begin{bmatrix} 
        O & \Invg_{t,xe} \\ \Sigma_w^{-1/2} & O \\
        O & \Invg_{t,ee}
    \end{bmatrix} \in \R^{(\dmu + \dx) \times \dmu} \\
    J_M &= \begin{bmatrix}
        \Invg_{t,xx} \\ -\Sigma_w^{-1/2} G_t \\
        \Invg_{t,ex}
    \end{bmatrix} \in \R^{(\dmu + \dx) \times \dx},
\end{align*}
where we have defined $G_t$ to be the Jacobian of $g: \R^{\dx} \ra \R^{\dx}$ at $\overline{\mu_{t,x}} \in \R^{\dx}$, i.e.:
\begin{align*}
    G_t := \frac{\partial g}{\partial \xrt} \Bigg|_{\xrt = \overline{\mu_{t,x}}}
\end{align*}
Applying the Marginalization equations, we thus have:
\begin{align*}
\mu_{t+1} &\gets \tilde{x}_{t,K} - \Sigma_{t+1} J_K^\top \big[ I - J_M(J_M^\top J_M)^{-1} J_M^\top \big] \\
&\hspace{1cm} C_{M}(\overline{{\xr_K}^\star}, \overline{{\xr_M}^\star}) \\
&= g(\overline{\mu_t}), \\
\Sigma_{t+1} &\gets \big(J_K^\top \big[ I - J_M(J_M^\top J_M)^{-1} J_M^\top \big] J_K\big)^{-1}, \\ 
&= \big(J_K^\top J_K - J_K^\top J_M(J_M^\top J_M)^{-1} J_M^\top J_K\big)^{-1}, \\ 
&= \Bigg(\begin{bmatrix}
\Sigma_w^{-1} & O \\
O & \Invg_{ex} \Invg_{xe} + \Invg_{ee}^2
\end{bmatrix} 
- \begin{bmatrix}
- \Sigma_w^{-1} G_t \\
\Invg_{ex} \Invg_{xx} + \Invg_{ee} \Invg_{ex}
\end{bmatrix} \\
&\hspace{1cm} (\Invg_{xx}^2 + \Invg_{xe} \Invg_{ex} + G_t^\top \Sigma_w^{-1} G_t)^{-1} \\
&\hspace{1cm} \cdot \begin{bmatrix}
-G_t^\top \Sigma_w^{-1} & \Invg_{xx} \Invg_{xe} + \Invg_{ex} \Invg_{ee}
\end{bmatrix} \Bigg)^{-1} \\
&= \Bigg(\begin{bmatrix}
        \Sigma_w^{-1} & O \\
        O & \Omega_{ee}
        \end{bmatrix} 
        - \begin{bmatrix}
        - \Sigma_w^{-1} G_t \\
        \Omega_{ex}
        \end{bmatrix} \\ 
        &\hspace{1cm}(\Omega_{xx} + G_t^\top \Sigma_w^{-1} G_t)^{-1} \begin{bmatrix}
        -G_t^\top \Sigma_w^{-1} & \Omega_{xe}
    \end{bmatrix} \Bigg)^{-1}
\end{align*}
To show that this is indeed identical to the propagation equation for the covariance matrix in the standard formulation of the dynamic Extended Kalman Filter algorithm, i.e. Algorithm \ref{Alg: Dynamic EKF, Standard}, Line \ref{Eqn: Dynamic EKF, Cov Propagation}, we must show that:
\begin{align*}
    &\Bigg(\begin{bmatrix}
        \Sigma_w^{-1} & O \\
        O & \Omega_{ee}
        \end{bmatrix} 
        - \begin{bmatrix}
        - \Sigma_w^{-1} G_t \\
        \Omega_{ex}
        \end{bmatrix} (\Omega_{xx} + G_t^\top \Sigma_w^{-1} G_t)^{-1} \\
        &\hspace{1cm}
        \begin{bmatrix}
        -G_t^\top \Sigma_w^{-1} & \Omega_{xe}
    \end{bmatrix} \Bigg)^{-1} \\
    = \hspace{0.5mm} & \begin{bmatrix}
            G_t \overline{\Sigma}_{t,xx} G_t^\top + \Sigma_w & G_t \overline{\Sigma}_{t,xe} \\
            \overline{\Sigma}_{t,xe} G_t^\top & \overline{\Sigma}_{t,ee}
        \end{bmatrix}
\end{align*}
This follows by brute-force expanding the above block matrix components, and applying Woodbury's Matrix Identity, along with the definitions of $\Sigma_{t,xx}, \Invg_{t,xx}$, $\Sigma_{t,xe}, \Invg_{t,xe}$, $\Sigma_{t,ex}, \Invg_{t,ex}$, $\Sigma_{t,ee}$, and $\Invg_{t,ee}$.
\end{proof}


\subsection{Auxiliary Theorems}
\label{subsec: A2, Auxiliary Theorems}

For the proof of Theorem \ref{Thm: Dynamic EKF, Feature Augmentation} above, we require the following result, an extension of the classic Implicit Function Theorem and of \cite{Dontchev2005ImplicitFunctionsAndSolutionMappings}, Theorem 1F.6. This result appears as Exercise 1F.9 in \cite{Dontchev2005ImplicitFunctionsAndSolutionMappings}, and is presented here, with proof, for completeness.

\begin{theorem} \label{Thm: Implicit Function Theorem, Extension}
Let $f: \R^n \times \R^d \ra \R^p$ be a $k$-times continuously differentiable (i.e., $C^k$) function, with $p \leq d$. Fix $x^\star \in \R^n$, $y^\star \in \R^d$ arbitrarily, and let $z^\star := f(x^\star, y^\star) \in \R^p$.    Define $df_x:= \frac{\partial f}{\partial x} \in \R^{p \times n}$ and $df_y := \frac{\partial f}{\partial y} \in \R^{p \times d}$ at each point in $\R^n \times \R^d$, the domain of $f$. If $df_y(x^\star, y^\star)$ is surjective, then there exists open neighborhoods $U_x \subset \R^n$, $U_y \subset \R^d$, and $U_z \subset \R^p$ of $x^\star$, $y^\star$, and $z^\star$, respectively, and a $C^k$ map $g: U_x \times U_z \ra U_y$, such that:
\begin{enumerate}
    \item $f \big( x, g(x, z) \big) = z$, $\forall \hspace{0.5mm} x \in U_z$, $z \in U_z$.
    
    \item $dg_x(x^\star, z^\star) = \big( - df_y^\top (df_y df_y^\top)^{-1} df_x (x^\star, y^\star)$ \big), where $dg_x := \frac{\partial g}{\partial x}$ in the domain of $g$.
\end{enumerate}
\end{theorem}

\begin{proof}
Define $\bar f: \R^n \times \R^k \ra \R^n \times \R^p$ by $\bar f(x, y) := \big(x, f(x, y) \big)$ for each $(x, y) \in \R^n \times \R^k$. Then $d \bar f_x := \frac{\partial \bar f}{\partial x}$ is surjective at $(x^\star, y^\star)$. Thus, by \cite{Dontchev2005ImplicitFunctionsAndSolutionMappings}, Theorem 1F.6 (an extension of the Inverse Function Theorem), there exist open neighborhoods $U_x \subset \R^n$, $U_y \subset \R^d$, and $U_z \subset \R^p$ of $x^\star$, $y^\star$, and $z^\star$, respectively, and a $C^k$ map $\overline{g}: U_x \times U_z \ra U_x \times U_y$, such that:
\begin{align*}
    &\bar f \big( \bar g(x, z) \big) = (x, z), \hspace{1cm} \forall \hspace{0.5mm} (x, y) \in U_x \times U_y, \\
    &d \bar g_x (x^\star, z^\star) = \big( - d \bar f_y^\top (d \bar f_y d \bar f_y^\top)^{-1} d \bar f_x \big) (x^\star, y^\star),
\end{align*}
where $d \bar g_x := \frac{\partial \bar g}{\partial x}$ in the domain of $\overline{g}$.

Let the components of $\overline{g}$ be given by $\overline{g}(x, z) := \big( g_1(x, z), g_2(x, z) \big)$ for each $x \in U_x$, $z \in U_z$, with $g_1: U_x \times U_z \ra U_x$ and $g_2: U_x \times U_z \ra U_y$. We claim that $g_2$ is our desired function. First, for each $x \in U_x$ and $z \in U_z$:
\begin{align*}
    (x, z) &= \bar f \big( \bar g(x, z) \big) = \bar f \big( g_1(x, z), g_2(x, z) \big) \\
    &= \big(g_1(x, z), f \big( g_1(x, z), g_2(x, z) \big) \big),
\end{align*}
so $g_1(x, z) = x$, and thus, as desired:
\begin{align*}
    (x, z) &= \big(x, f \big(x, g_2(x, z) \big) \big)
\end{align*}
Next, let $df \in \R^{(n+p) \times k}$, $d \bar f \in \R^{(n+p) \times (n+k)}$, and $d \bar g \in \R^{(n+k) \times (n+p)}$ denote the (full) derivatives of the maps $f$, $\bar f$, and $\bar g$, in their respective domains. Then:
\begin{align*}
    &\hspace{5mm} d \bar g(x^\star, z^\star) \\
    &= \big( - d \bar f_y^\top (d \bar f_y d \bar f_y^\top)^{-1} d \bar f_x \big) (x^\star, y^\star) \\
    &= \begin{bmatrix}
        I_{n \times n} & df_x^\top \\
        O_{k \times n} & df_y^\top
    \end{bmatrix}
    \left( \begin{bmatrix}
        I_{n \times n} & O_{k \times n} \\
        df_x & df_y
    \end{bmatrix}
    \begin{bmatrix}
        I_{n \times n} & df_x^\top \\
        O_{k \times n} & df_y^\top
    \end{bmatrix} \right)^{-1} \\
    &\hspace{1cm} (x^\star, y^\star)  \\
    &= \begin{bmatrix}
        I_{n \times n} & df_x^\top \\
        O_{k \times n} & df_y^\top
    \end{bmatrix}
    \begin{bmatrix}
        I_{n \times n} & df_x^\top \\
        df_x & df_x df_x^\top + df_y df_y^\top
    \end{bmatrix}^{-1} (x^\star, y^\star) \\
    &= \begin{bmatrix}
        I_{n \times n} & O_{k \times n} \\
        - df_y^\top (df_y df_y^\top)^{-1} df_x & df_y^\top (df_y df_y^\top)^{-1}
    \end{bmatrix} (x^\star, y^\star).
\end{align*}
Consequently:
\begin{align*}
    d \overline (g_2)_x (x^\star, z^\star) = - df_y^\top (df_y df_y^\top)^{-1} df_x (x^\star, y^\star),
\end{align*}
as desired.
\end{proof}

For the proof of Theorem \ref{Thm: Dynamic EKF, Dynamic Object Pose Augmentation} above, we require the following results, derived from an corollary to the Global Rank Theorem (\cite{Lee2000IntroductionToSmoothManifolds}, Theorem 4.14).

\begin{theorem}[Corollary to \cite{Lee2000IntroductionToSmoothManifolds}, Theorem 4.14]
Let $\mcalM$ and $\mcalN$ denote manifolds of dimension $d_\mcalM$ and $d_\mcalN$, respectively, with $d_\mcalM \leq d_\mcalN$, and let $f: \mcalM \ra \mcalN$ be a smooth local injection about the point $p \in \mcalM$. Then for any pair of smooth local charts $(U_\mcalM, \phi_\mcalM)$, $(U_\mcalN, \phi_\mcalN)$ centered at $p$ and $f(p)$, respectively, where $U_\mcalM$, $U_\mcalN$ are open neighborhoods of $p$ and $f(p)$ in $\mcalM$ and $\mcalN$, respectively, with $f(U_\mcalM) \subset U_\mcalN$, there exists:
\begin{enumerate}
    \item Open neighborhoods $U_\mcalM' \subset U_\mcalM$ and $U_\mcalN' \subset U_\mcalN$ of $p$ and $f(p)$ in $\mcalM$ and $\mcalN$, respectively, and
    
    \item A local inverse map $g: U_\mcalN' \ra U_\mcalM'$, such that $(g \circ f)(p') = p'$ for all $p' \in \mcalM'$, and:
    \begin{align*}
        \frac{d}{dy^i} \big( \phi_\mcalM \circ g \circ \phi_\mcalN^{-1} \big) = \Bigg( \frac{d}{dx^i} \big( \phi_\mcalN \circ f \circ \phi_\mcalM^{-1} \big) \Bigg)^\dagger,
    \end{align*}
    where $(x^1, \cdots, x^m)$ and $(y^1, \cdots, y^n)$ denote Euclidean coordinates in $\phi_\mcalM(U_\mcalM')$ and $\phi_\mcalN(U_\mcalN')$, respectively, and $\dagger$ denotes the Moore-Penrose pseudoinverse.
\end{enumerate}
\end{theorem}

\begin{proof}
By the Global Rank Theorem (\cite{Lee2000IntroductionToSmoothManifolds}, Theorem 4.14), there exists a pair of smooth local charts, $(\overline U_\mcalM, \overline \phi_\mcalM)$ and $(\overline U_\mcalN, \overline \phi_\mcalN)$, centered at $p$ and $f(p)$, respectively, such that, for any $x := (x_1, \cdots, x_m) \in \overline \phi_\mcalM(\overline U_\mcalM)$:
\begin{align*}
    (\overline \phi_\mcalN \circ f \circ \overline \phi_\mcalM^{-1})(x_1, \cdots, x_m) = (x_1, \cdots, x_m, 0, \cdots, 0).
\end{align*}
Define $\proj: \overline \phi_\mcalN(\overline U_\mcalN) \ra \overline \phi_\mcalM(\overline U_\mcalM)$
by:
\begin{align*}
    \embed(x) &:= (x_1, \cdots, x_m, 0, \cdots, 0)
    \proj(y) &:= (y_1, \cdots, y_m), 
\end{align*}
for each $x := (x_1, \cdots, x_m) \in \phi_\mcalN(\overline U_\mcalN)$ and $y := (y_1, \cdots, y_m, \cdots, y_n) \in \phi_\mcalM(\overline U_\mcalM)$. 

Now, observe that:
\begin{align*}
    &\phi_\mcalN \circ f \circ \phi_\mcalM^{-1} \\
    = \hspace{0.5mm} &(\phi_\mcalN \circ \overline \phi_\mcalN^{-1}) \circ (\overline \phi_\mcalN \circ f \circ \overline \phi_\mcalM^{-1}) \circ (\overline \phi_\mcalM \circ \phi_\mcalM^{-1}), \\
    \Ra \hspace{5mm} &d(\phi_\mcalN \circ f \circ \phi_\mcalM^{-1}) \\
    = \hspace{0.5mm} &d(\phi_\mcalN \circ \overline \phi_\mcalN^{-1}) \circ d(\overline \phi_\mcalN \circ f \circ \overline \phi_\mcalM^{-1}) \circ d(\overline \phi_\mcalM \circ \phi_\mcalM^{-1}) \\
    = \hspace{0.5mm} &d(\phi_\mcalN \circ \overline \phi_\mcalN^{-1}) \circ \begin{bmatrix}
        I_{m \times m} \\ O_{(n-m) \times m}
    \end{bmatrix} \circ d(\overline \phi_\mcalM \circ \phi_\mcalM^{-1}),
\end{align*}
where $d(\cdot)$ denotes the Jacobian of a smooth map.
Intuitively, we wish to take the pseudoinverse of the matrix near the \say{center} of the above expression. Since $d(\phi_\mcalN \circ \overline \phi_\mcalN^{-1})$ and $d(\overline \phi_\mcalM \circ \phi_\mcalM^{-1})$ are, in general, not orthogonal matrices, some additional processing is required. In particular, let $Q_\mcalN R_\mcalN$ and $Q_\mcalM R_\mcalM$ be the QR decomposition of $d(\phi_\mcalN \circ \overline \phi_\mcalN^{-1})$ and $d(\phi_\mcalM \circ \overline \phi_\mcalM^{-1})$, respectively, and define $g: U_\mcalN \cap \overline{U}_\mcalN \ra U_\mcalM \cap \overline{U}_\mcalM$ by:
\begin{align*}
    g &:= \overline \phi_\mcalM^{-1} \circ L_{R_\mcalM^{-1}} \circ L_{\big( L_{R_\mcalN} \circ \overline \phi_\mcalN \circ f \circ \overline \phi_\mcalM^{-1} \circ L_{R_\mcalM}^{-1} \big)^\dagger} \\
    &\hspace{1cm} \circ L_{R_\mcalN} \circ \overline \phi_\mcalN
\end{align*}
where, given any matrix $A \in \R^{m \times n}$, we use $L_A: \R^n \times \R^m$ to denote the corresponding linear map. Then:
\begin{align*}
    &(g \circ f)(p) \\
    = \hspace{0.5mm} &\big( \overline \phi_\mcalM^{-1} \circ L_{R_\mcalM^{-1}} \circ \big( L_{R_\mcalN} \circ \overline \phi_\mcalN \circ f \circ \overline \phi_\mcalM^{-1} \circ L_{R_\mcalM}^{-1} \big) \\
    &\hspace{5mm} \circ L_{R_\mcalN} \circ \overline \phi_\mcalN \big) \\
    &\hspace{5mm} \circ \big( \overline \phi_\mcalN^{-1} \circ L_{R_\mcalN^{-1}} \circ L_{R_\mcalN} \circ \overline \phi_\mcalN \circ f \\
    &\hspace{1cm} \circ \overline \phi_\mcalM^{-1} \circ L_{R_\mcalM^{-1}} \circ L_{R_\mcalM} \circ \overline \phi_\mcalM \big)(p) \\
    = \hspace{0.5mm} &p,
\end{align*}
for any $p \in \mcalM$, and:
\begin{align*}
    &\left( \frac{d}{dx} \big( \phi_\mcalN \circ f \circ \phi_\mcalM^{-1} \big) \right)^\dagger \\
    = \hspace{0.5mm} &\left( L_{Q_\mcalN} \circ L_{R_\mcalN} \circ \embed \circ L_{R_\mcalM^{-1}} \circ L_{Q_\mcalM^{-1}} \right)^\dagger \\
    = \hspace{0.5mm} &Q_\mcalM \left( R_\mcalN \begin{bmatrix}
        I_{m \times m} & O_{m \times (n-m)}
    \end{bmatrix} R_\mcalM^{-1} \right)^\dagger Q_\mcalN^{-1},
\end{align*} 
whereas:
\begin{align*}
    &\frac{d}{dy}(\phi_\mcalM \circ g \circ \phi_{\mcalN}^{-1}) \\
    = \hspace{0.5mm} &\Bigg( \phi_\mcalM \circ \phi_\mcalM^{-1} \circ L_{R_\mcalM^{-1}} \\
    &\hspace{5mm} \circ L_{\big( L_{R_\mcalN} \circ \overline \phi_\mcalN \circ f \circ \overline \phi_\mcalM^{-1} \circ L_{R_\mcalM}^{-1} \big)^\dagger} \circ L_{R_\mcalN} \\
    &\hspace{5mm} \circ \overline \phi_\mcalN \circ \phi_{\mcalN}^{-1} \Bigg) \\
    = \hspace{0.5mm} &Q_\mcalM R_\mcalM \\
    &\hspace{5mm} \cdot R_\mcalM \Bigg(R_\mcalN \cdot \begin{bmatrix}
        I_{m \times m} & O_{m \times (n-m)}
    \end{bmatrix} \cdot R_\mcalM^{-1} \Bigg)^\dagger R_\mcalN \\
    &\hspace{5mm} \cdot R_\mcalN^{-1} Q_\mcalN^{-1} \\
    = \hspace{0.5mm} &Q_\mcalM \left( R_\mcalN \begin{bmatrix}
        I_{m \times m} & O_{m \times (n-m)}
    \end{bmatrix} R_\mcalM^{-1} \right)^\dagger Q_\mcalN^{-1},
\end{align*}
so we have:
\begin{align*}
    \frac{d}{dy^i} \big( \phi_\mcalM \circ g \circ \phi_\mcalN^{-1} \big) = \Bigg( \frac{d}{dx^i} \big( \phi_\mcalN \circ f \circ \phi_\mcalM^{-1} \big) \Bigg)^\dagger,
\end{align*}
as claimed.

\end{proof}

\subsection{Experiment Details}
\label{subsec: A3, Experiment Details}

Additional details regarding the experiment settings for the simulation described in Section \ref{sec: Experiments} are as follows:
\begin{itemize}
    \item The dynamics map used is $g: \R^3 \ra \R^3$, as given by:
    \begin{align*}
        g(x) := x + x_{\text{odom}} + w, \hspace{5mm} w \sim \mathcal{N}(0, \Sigma_w),
    \end{align*}
    for each $x, x_{\text{odom}} \in \R^3$.
    with $\Sigma_w \in \R^{3 \times 3}$ as one of the three choices given in \ref{Eqn: Sigma w, v, 1}, \ref{Eqn: Sigma w, v, 2}, and \ref{Eqn: Sigma w, v, 3}.
    
    \item The measurement map used is $h: \R^3 \times \R^2 \ra \R^2$, given by:
    \begin{align*}
        h(x, f) &:= \begin{bmatrix}
            \cos x_3 & \sin x_3 \\
            - \sin x_3 & \cos x_3
        \end{bmatrix}
        \begin{bmatrix}
            f_1 - x_1 \\
            f_2 - x_2
        \end{bmatrix}
        + v, \\
        &\hspace{5mm} v \sim \mathcal{N}(0, \Sigma_v),
    \end{align*}
    for each $x := (x_1, x_2, x_3) \in \R^3$ and $f := (f_1, f_2) \in \R^2$,
    with $\Sigma_v \in \R^{2 \times 2}$ as one of the three choices given in \ref{Eqn: Sigma w, v, 1}, \ref{Eqn: Sigma w, v, 2}, and \ref{Eqn: Sigma w, v, 3}.
    
    \item The moving object pose map transform $g^o: \R^3 \times \R^{2 \nof} \ra \R^{2 \nof}$ is given by:
    \begin{align*}
        &f^c := \frac{1}{\nof} \sum_{j=1}^{\nof} f_j, \\
        &g^o(\fpose, f_{\text{vec}}) \\
        := \hspace{0.5mm} &\text{vec} \Bigg( 
        \begin{bmatrix}
            \cos \fpose_3 & - \sin \fpose_3 \\
            \sin \fpose_3 & \cos \fpose_3
        \end{bmatrix} \\
        &\hspace{1cm} \cdot \begin{bmatrix}
            f_1^1 - f_1^c & \cdots & f_1^n - f_1^c \\
            f_2^1 - f_2^c & \cdots & f_2^n - f_2^c
        \end{bmatrix} \\
        &\hspace{5mm} +
        \begin{bmatrix}
            f_1^c + \fpose_1 \\
            f_2^c + \fpose_2
        \end{bmatrix}
        \begin{bmatrix}
            1 & \cdots & 1
        \end{bmatrix} \Bigg), \\
        = \hspace{0.5mm} &\begin{bmatrix}
            (f_1^1 - f_1^c) \cos \fpose_3 -  (f_2^1 - f_2^c) \sin \fpose_3 + f_1^c + \fpose_1 \\
            (f_2^1 - f_2^c) \cos \fpose_3 -  (f_2^1 - f_2^c) \sin \fpose_3 + f_2^c + \fpose_2 \\
            \vdots \\
            (f_1^n - f_1^c) \cos \fpose_3 -  (f_2^n - f_2^c) \sin \fpose_3 + f_1^c + \fpose_1 \\
            (f_2^n - f_2^c) \cos \fpose_3 -  (f_2^n - f_2^c) \sin \fpose_3 + f_2^c + \fpose_2
        \end{bmatrix}
    \end{align*}
    Here, we have defined $f_{\text{vec}} := (f_1^1, f_2^1, \cdots, f_1^n, f_2^n) \in \R^{2n}$, and defined the \textit{vectorizaton operator} $\text{vec}$ as follows: Given $A \in \R^{m \times n}$, we have $\text{vec}(A) := (A_{11}, \cdots, A_{1n}, \cdots, A_{m1}, \cdots, A_{mn}) \in \R^{mn}$.
    
    \item For the moving pose augmentation step, we use $\Sigma_{\fpose} := 0.1 \cdot I_{2 \times 2}$ for all simulations. The SVD solution to Wahba's problem is used to initialize all pose transformations from positions of each moving object feature observed at the initial time and at the current time.
    
    \item Due to the relatively smooth trajectories of this simulated scenario, the smoothing cost was omitted.
    
    \item As discussed in the main body of this tutorial, past moving object features and poses are dropped from the optimization window to ease the computational burden of the dynamic EKF SLAM algorithm at future timesteps. Note that, in real-life autonomous navigation tasks, dropping such poses and features far in the past would not severely impact the pose and feature estimation capabilities of the ego robot in the present and near future.
\end{itemize}





\end{document}